\documentclass[10pt,twocolumn,letterpaper]{article}

\usepackage{cvpr}
\usepackage{times}
\usepackage{epsfig}
\usepackage{graphicx}
\usepackage{amsmath}
\usepackage{amssymb}
\usepackage{units}
\usepackage{subfigure}
\usepackage{algpseudocode}
\usepackage{algorithm}
\usepackage{mathtools}
\usepackage{bbm}
\usepackage{xfrac}
\usepackage{enumitem}
\usepackage{multirow}
\usepackage{colortbl}
\usepackage{textcomp}
\usepackage{fancyvrb}
\usepackage{url}
\usepackage{relsize}
\usepackage{appendix}

\allowdisplaybreaks

\DeclarePairedDelimiter{\abs}{\lvert}{\rvert}
\DeclarePairedDelimiter{\norm}{\lVert}{\rVert}

\newcommand{\power}{\alpha}

\usepackage{expl3}
\ExplSyntaxOn
\newcommand\latinabbrev[1]{
  \peek_meaning:NTF . {
    #1\@}%
  { \peek_catcode:NTF a {
      #1.\@ }%
    {#1.\@}}}
\ExplSyntaxOff

\usepackage[breaklinks=true,bookmarks=false]{hyperref}

\cvprfinalcopy

\def\cvprPaperID{1472}

\ifcvprfinal\fi

\begin{document}

\newcommand{\charb}[1]{d\left( #1 \right)}
\newcommand{\gencharb}[1]{g\left( #1 \right)}
\newcommand{\rawlossfun}[1]{f\left( #1 \right)}
\newcommand{\lossfun}[1]{ \rho\left( #1 \right)}
\newcommand{\deriv}{\partial}
\newcommand{\prob}[1]{p\left( #1 \right)}
\newcommand{\partition}[1]{Z\left( #1 \right)}
\newcommand{\entropy}[1]{H\left( #1 \right)}

\title{A General and Adaptive Robust Loss Function}

\author{
Jonathan T. Barron
\\
Google Research
}

\maketitle

\begin{abstract}

We present a generalization of the
Cauchy/Lorentzian, Geman-McClure, Welsch/Leclerc, generalized Charbonnier,
Charbonnier/pseudo-Huber/L1-L2, and L2 loss functions.
By introducing robustness as a continuous parameter, our loss function allows algorithms
built around robust loss minimization to be generalized,
which improves performance on basic vision tasks such as registration and clustering.
Interpreting our loss as the negative log of a univariate density yields a general
probability distribution that includes normal and Cauchy distributions as special cases.
This probabilistic interpretation enables the training of neural networks in which the robustness
of the loss automatically adapts itself during training,
which improves performance on learning-based tasks such as
generative image synthesis and unsupervised monocular depth estimation, without
requiring any manual parameter tuning.

\end{abstract}

Many problems in statistics and optimization require \emph{robustness} ---
that a model be less influenced by outliers than by inliers \cite{Hastie_et_al_2015,huber1981}.
This idea is common in parameter estimation and learning tasks, where a robust
loss (say, absolute error) may be preferred over a non-robust loss
(say, squared error) due to its reduced sensitivity to large errors.
Researchers have developed various robust penalties with particular properties,
many of which are summarized well in \cite{BlackR96, Zhang95parameterestimation}.
In gradient descent or M-estimation \cite{HampelEtal86} these losses are often interchangeable,
so researchers may experiment with different losses when designing a system.
This flexibility in shaping a loss function may be useful because of non-Gaussian noise,
or simply because the loss that is minimized during learning or parameter estimation is
different from how the resulting learned model or estimated parameters will be evaluated.
For example, one might train a neural network by minimizing the difference between the
network's output and a set of images, but evaluate that network
in terms of how well it hallucinates random images.

In this paper we present a single loss function that is a superset of many common
robust loss functions.
A single continuous-valued parameter in our general loss function can be set such that
it is equal to several traditional losses, and
can be adjusted to model a wider family of functions.
This allows us to generalize algorithms built around a fixed robust loss
with a new ``robustness'' hyperparameter that can be tuned or annealed to improve performance.

\begin{figure}[b]
\centering
\begin{tabular}{@{}c@{}c@{}}
  \includegraphics[width=0.5\linewidth]{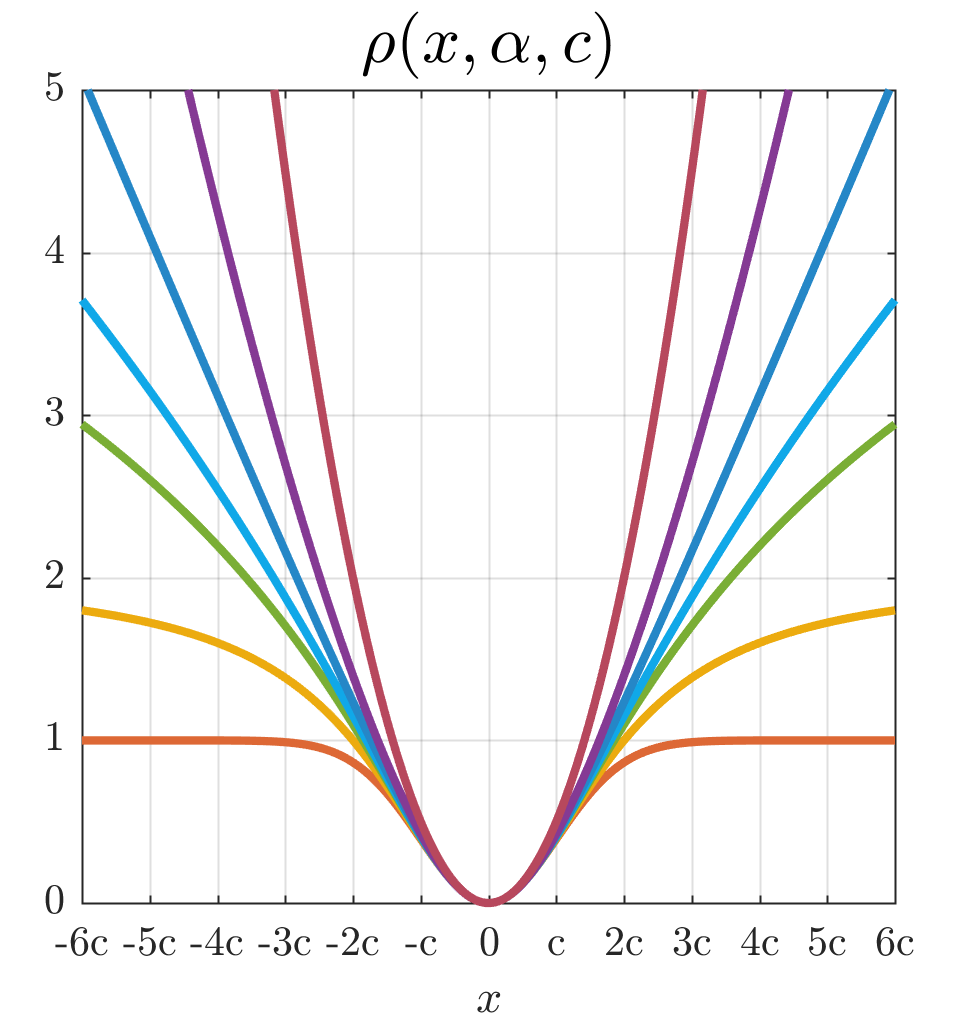} &
  \includegraphics[width=0.5\linewidth]{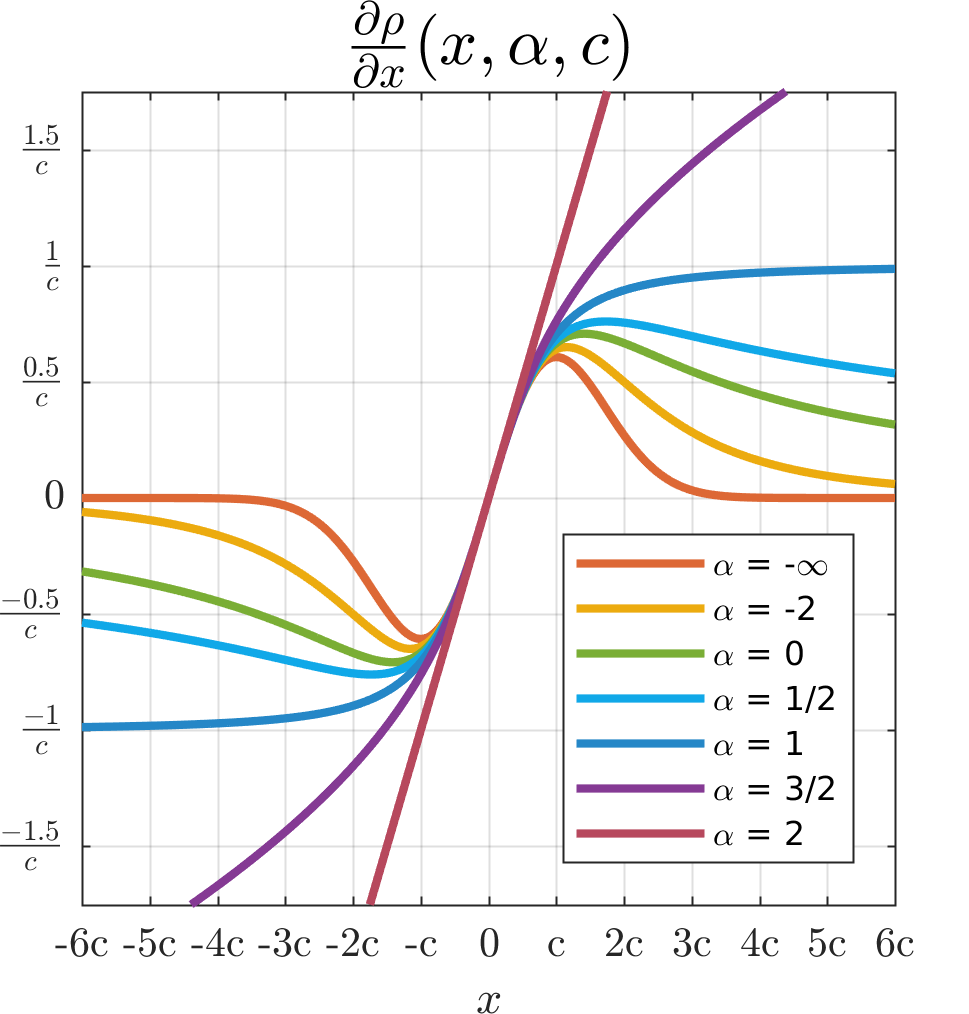}
\end{tabular}
  \caption{
   Our general loss function (left) and its gradient (right)
   for different values of its shape parameter $\power$.
   Several values of $\power$ reproduce existing loss functions: L2 loss ($\power=2$),
   Charbonnier loss ($\power=1$), Cauchy loss ($\power=0$), Geman-McClure loss ($\power=-2$),
   and Welsch loss ($\power=-\infty$).
   \label{fig:loss_and_grad}
  }
\end{figure}

Though new hyperparameters may be valuable to a practitioner, they complicate
experimentation by requiring manual tuning or time-consuming cross-validation.
However, by viewing our general loss function as the negative
log-likelihood of a probability distribution, and by treating the robustness of that distribution
as a latent variable, we show that maximizing the likelihood of that distribution
allows gradient-based optimization frameworks to \emph{automatically}
determine how robust the loss should be without any manual parameter tuning.
This ``adaptive'' form of our loss is particularly effective in models with
multivariate output spaces (say, image generation or depth estimation)
as we can introduce independent robustness variables for each dimension
in the output and thereby allow the model to independently adapt the robustness
of its loss in each dimension.

The rest of the paper is as follows:
In Section~\ref{sec:loss} we define
our general loss function, relate it to existing losses, and enumerate
some of its useful properties.
In Section~\ref{sec:pdf} we use our loss to construct a probability
distribution, which requires deriving a partition function and a sampling procedure.
Section~\ref{sec:experiments} discusses four representative experiments:
In Sections~\ref{sec:vae} and ~\ref{sec:monodepth}
we take two vision-oriented deep learning models (variational autoencoders for image synthesis and self-supervised monocular depth estimation),
replace their losses with the negative log-likelihood of our general distribution,
and demonstrate that allowing our distribution to automatically determine its own
robustness can improve performance without introducing any additional manually-tuned hyperparameters.
In Sections~\ref{sec:registration} and \ref{sec:rcc} we use our loss function
to generalize algorithms for the classic vision tasks of registration and clustering,
and demonstrate the performance improvement that can be achieved by introducing
robustness as a hyperparameter that is annealed or manually tuned.

\section{Loss Function}
\label{sec:loss}

The simplest form of our loss function is:
\begin{equation}
\rawlossfun{x, \power, c} = {\abs{\power - 2} \over \power } \left( \left( {\left( \sfrac{x}{c} \right)^2 \over \abs{\power - 2} } + 1 \right)^{\sfrac{\power}{2}} - 1 \right)
\end{equation}
Here $\power \in \mathbb{R}$ is a shape parameter that controls the robustness of the loss
and $c > 0$ is a scale parameter that controls the size of the loss's quadratic bowl near $x=0$.

Though our loss is undefined when $\power = 2$, it approaches L2 loss (squared error) in the limit:
\begin{equation}
\lim_{\power \to 2} \rawlossfun{x, \power, c } = {1 \over 2} \left( \sfrac{x}{c} \right)^2
\end{equation}
When $\power=1$ our loss is a smoothed form of L1 loss:
\begin{equation}
\rawlossfun{x, 1, c } = \sqrt{(\sfrac{x}{c})^2 + 1} - 1
\end{equation}
This is often referred to as Charbonnier loss \cite{Charb1994},
pseudo-Huber loss (as it resembles Huber loss \cite{huber1964}),
or L1-L2 loss \cite{Zhang95parameterestimation} (as it behaves like L2 loss near the origin and like L1 loss elsewhere).

Our loss's ability to express L2 and smoothed L1 losses is shared by the ``generalized Charbonnier'' loss \cite{SunCVPR10},
which has been used in flow and depth estimation tasks that require robustness \cite{chen2014fast, Krahenbuhl2012} and is
commonly defined as:
\begin{equation}
\left(x^2 + \epsilon^2 \right)^{\sfrac{\power}{2}}
\end{equation}
Our loss has significantly more expressive power than the generalized
Charbonnier loss, which we can see by setting our shape parameter $\power$ to
nonpositive values. Though $\rawlossfun{x, 0, c}$ is undefined, we can take
the limit of $\rawlossfun{x, \power, c}$ as $\power$ approaches zero:
\begin{equation}
\lim_{\power \to 0} \rawlossfun{x, \power, c} = \log \left( {1 \over 2} \left( \sfrac{x}{c} \right)^2 + 1 \right)
\end{equation}
This yields Cauchy (aka Lorentzian) loss \cite{black1996robust}.
By setting $\power = -2$, our loss reproduces Geman-McClure loss \cite{geman1985}:
\begin{equation}
\rawlossfun{x, -2, c } = { 2 \left( \sfrac{x}{c} \right)^2 \over \left( \sfrac{x}{c} \right)^2 + 4}
\end{equation}
In the limit as $\power$ approaches negative infinity, our loss becomes Welsch \cite{DenWel1978} (aka Leclerc \cite{leclerc1989constructing}) loss:
\begin{equation}
\lim _{\power \to -\infty} \rawlossfun{x, \power, c } = 1 - \exp \left( - {1 \over 2} \left( \sfrac{x}{c} \right)^2 \right)
\end{equation}
With this analysis we can present our final loss function, which is simply $\rawlossfun{\cdot}$
with special cases for its removable singularities at $\power=0$ and $\power=2$ and its limit at $\power=-\infty$.
\begin{equation}
\resizebox{\linewidth}{!}{$%
\lossfun{x, \power, c} = \begin{cases}
{1 \over 2} \left( \sfrac{x}{c} \right)^2 & \text{if } \power = 2 \\
\log \left( {1 \over 2} \left( \sfrac{x}{c} \right)^2 + 1 \right) & \text{if } \power = 0 \\
1 - \exp \left( - {1 \over 2} \left( \sfrac{x}{c} \right)^2 \right) & \text{if } \power = -\infty \\
{\abs{\power - 2} \over \power } \left( \left( {\left( \sfrac{x}{c} \right)^2 \over \abs{\power - 2} } + 1 \right)^{\sfrac{\power}{2}} - 1 \right) & \text{otherwise}
\end{cases}
$}%
\end{equation}
As we have shown, this loss function is a superset of the
Welsch/Leclerc,
Geman-McClure,
Cauchy/Lorentzian,
generalized Charbonnier,
Charbonnier/pseudo-Huber/L1-L2, and
L2 loss functions.

To enable gradient-based optimization we can derive the derivative of $\lossfun{x, \power, c}$ with respect to $x$:
\begin{equation}
{ \deriv \rho \over \deriv x } \left(x, \power, c\right) = \begin{cases}
{x \over c^2} & \text{if } \power = 2 \\
{2x \over x^2 + 2c^2 } & \text{if } \power = 0 \\
{x \over c^2} \exp \left(- {1 \over 2} \left( \sfrac{x}{c} \right)^2 \right) & \text{if } \power = -\infty \\
{x \over c^2} \left( { \left( \sfrac{x}{c} \right)^2 \over \abs{\power - 2} } + 1 \right)^{(\sfrac{\power}{2} - 1)} & \text{otherwise}
\end{cases}
\end{equation}
Our loss and its derivative are visualized for different values of $\power$ in Figure~\ref{fig:loss_and_grad}.

The shape of the derivative gives some intuition as to how $\power$ affects behavior when our loss is being minimized by gradient descent
or some related method.
For all values of $\power$ the derivative is approximately linear when $\abs{x} < c$,
so the effect of a small residual is always linearly proportional to that residual's magnitude.
If $\power=2$, the derivative's magnitude stays linearly proportional to the residual's magnitude --- a larger residual has a
correspondingly larger effect.
If $\power=1$ the derivative's magnitude saturates to a constant $\sfrac{1}{c}$ as $\abs{x}$ grows larger than $c$, so
as a residual increases its effect never decreases but never exceeds a fixed amount.
If $\power < 1$ the derivative's magnitude begins to decrease as $\abs{x}$ grows larger than $c$
(in the language of M-estimation \cite{HampelEtal86}, the derivative, aka ``influence'', is ``redescending'')
so as the residual of an outlier increases, that outlier has \emph{less} effect during gradient descent.
The effect of an outlier diminishes as $\power$ becomes more negative, and as $\power$
approaches $-\infty$ an outlier whose residual magnitude is larger than $3c$
is almost completely ignored.

We can also reason about $\power$ in terms of averages.
Because the empirical mean of a set of values minimizes total squared error between the mean and the set,
and the empirical median similarly minimizes absolute error,
minimizing our loss with $\power=2$ is equivalent to estimating
a mean, and with $\power=1$ is similar to estimating a median.
Minimizing our loss with $\power=-\infty$ is equivalent to local mode-finding \cite{Boomgaard2002}.
Values of $\power$ between these extents can be thought of as smoothly interpolating between
these three kinds of averages during estimation.

Our loss function has several useful properties that we will take
advantage of.
The loss is smooth (\ie, in $C^\infty$) with respect to $x$, $\power$, and $c>0$,
and is therefore well-suited to gradient-based optimization over its input and
its parameters.
The loss is zero at the origin, and increases monotonically with respect to $\abs{x}$:
\begin{equation}
\lossfun{0, \power, c} = 0 \quad\quad { \deriv \rho \over \deriv \abs{x} } \left(x, \power, c\right) \geq 0
\end{equation}
The loss is invariant to a simultaneous scaling of $c$ and $x$:
\begin{equation}
\forall_{k>0} \,\,\, \rho(k x, \power, k c) = \rho(x, \power, c)\
\end{equation}
The loss increases monotonically with respect to $\power$:
\begin{equation}
{ \deriv \rho \over \deriv \power } \left(x, \power, c\right) \geq 0  \label{eq:monopower}
\end{equation}
This is convenient for graduated non-convexity~\cite{Blake1987}:
we can initialize $\power$ such that our loss is convex and then gradually reduce
$\power$ (and therefore reduce convexity and increase robustness) during optimization,
thereby enabling robust estimation that (often) avoids local minima.

We can take the limit of the loss as $\power$ approaches infinity,
which due to Eq.~\ref{eq:monopower} must be the upper bound of the loss:
\begin{equation}
\lossfun{x, \power, c} \leq \lim_{\power \to +\infty} \lossfun{x, \power, c} = \exp \left( {1 \over 2} \left( \sfrac{x}{c} \right)^2 \right) - 1 \label{eq:inf_power}
\end{equation}
We can bound the magnitude of the gradient of the loss, which allows us to better reason about exploding gradients:
\begin{equation}
\left\lvert { \deriv \rho \over \deriv x } \left(x, \power, c\right) \right\rvert \leq
\begin{cases}
{1 \over c} \left( { \power - 2 \over \power - 1 } \right)^{ \left( \power - 1 \over 2 \right) }  \leq {1 \over c} \quad & \text{if } \power \leq 1 \\
{\abs{x} \over c^2} \quad & \text{if } \power \leq 2
\end{cases}
\label{eq:gradbound}
\end{equation}
L1 loss is not expressible by our loss,
but if $c$ is much smaller than $x$ we can approximate it with $\power = 1$:
\begin{equation}
\rawlossfun{x, 1, c } \approx {\abs{x} \over c} - 1  \quad \text{if } c \ll x \label{eq:L1loss}
\end{equation}
See Appendix~\ref{app:additional_properties} for other potentially-useful properties that
are not used in our experiments.

\section{Probability Density Function}
\label{sec:pdf}

With our loss function we can construct a general probability distribution,
such that the negative log-likelihood (NLL) of its PDF is a shifted version of our loss function:
\begin{align}
\prob{x \;|\; \mu, \power, c} &= {1 \over c \partition{\power}} \exp \left( -\lossfun{x - \mu, \power, c} \right) \\
\partition{\power} &= \int_{-\infty}^{\infty} \exp \left( -\lossfun{x, \power, 1} \right)
\end{align}
where $\prob{x \;|\; \mu, \power, c}$ is only defined if $\power \geq 0$, as $\partition{\power}$ is divergent when
$\power < 0$. For some values of $\power$ the partition function is relatively straightforward:
\begin{align}
\partition{0} &= \pi \sqrt{2} & \partition{1} &= 2 e K_1(1) \nonumber \\
\partition{2} &= \sqrt{2 \pi} & \partition{4} &= e^{\sfrac{1}{4}} K_{\sfrac{1}{4}}(\sfrac{1}{4})
\end{align}
where $K_n(\cdot)$ is the modified Bessel function of the second kind.
For any rational positive $\power$ (excluding a singularity at $\power=2$) where $\power = \sfrac{n}{d}$ with $n, d \in \mathbb{N}$, we see that
\begin{equation}
\partition{\frac{n}{d}} = \frac{e^{\abs*{ \frac{2 d}{n} - 1}} \sqrt{\abs*{\frac{2 d}{n} - 1}}}{ (2 \pi)^{(d - 1)} } G_{p,q}^{\,0,0}\!\left(\!\left.\!
\begin{array}{c}
 \mathbf{a_p} \\
 \mathbf{b_q} \\
\end{array}
\right|
\left(\frac{1}{n} - \frac{1}{2d}\right)^{2d}
\right) \nonumber
\end{equation}
\vspace{-0.08in}
\begin{equation}
\resizebox{\linewidth}{!}{$%
\displaystyle
\mathbf{b_q} = \left\{ \left. \frac{i}{n} \right| i = -\frac{1}{2}, ... , n-\frac{3}{2} \right\} \cup \left\{ \left. \frac{i}{2 d} \right| i = 1, ... , 2d-1 \right\}  \nonumber
$}
\end{equation}
\begin{equation}
\mathbf{a_p} = \left\{ \left. \frac{i}{n} \right| i = 1, ... , n-1 \right\} \hspace{1.3in}
\end{equation}
where $G(\cdot)$ is the Meijer G-function and $\mathbf{b_q}$ is a multiset (items may occur twice).
Because the partition function is difficult to evaluate or differentiate,
in our experiments we approximate $\log(\partition{\power})$
with a cubic hermite spline (see Appendix~\ref{app:partition} for details).

\begin{figure}[t!]
\centering
  \begin{tabular}{@{}c@{}c@{}}
    \includegraphics[width=0.5\linewidth]{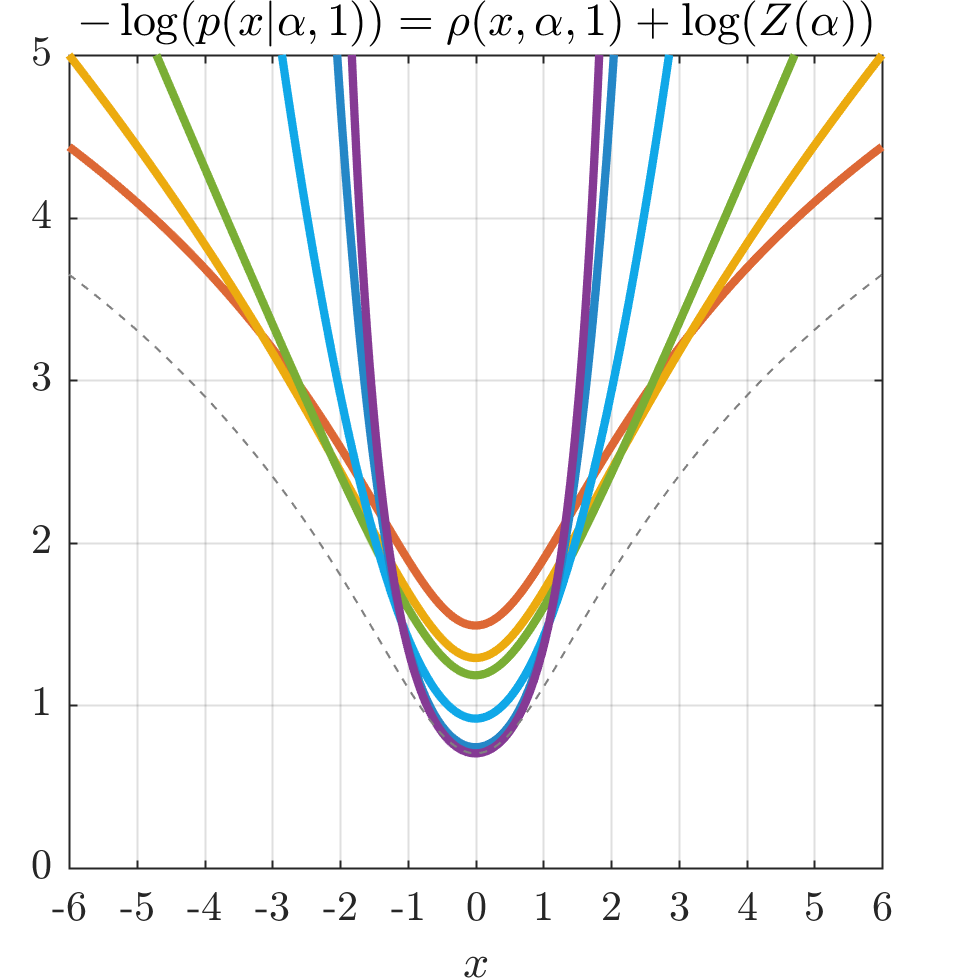} &
    \includegraphics[width=0.5\linewidth]{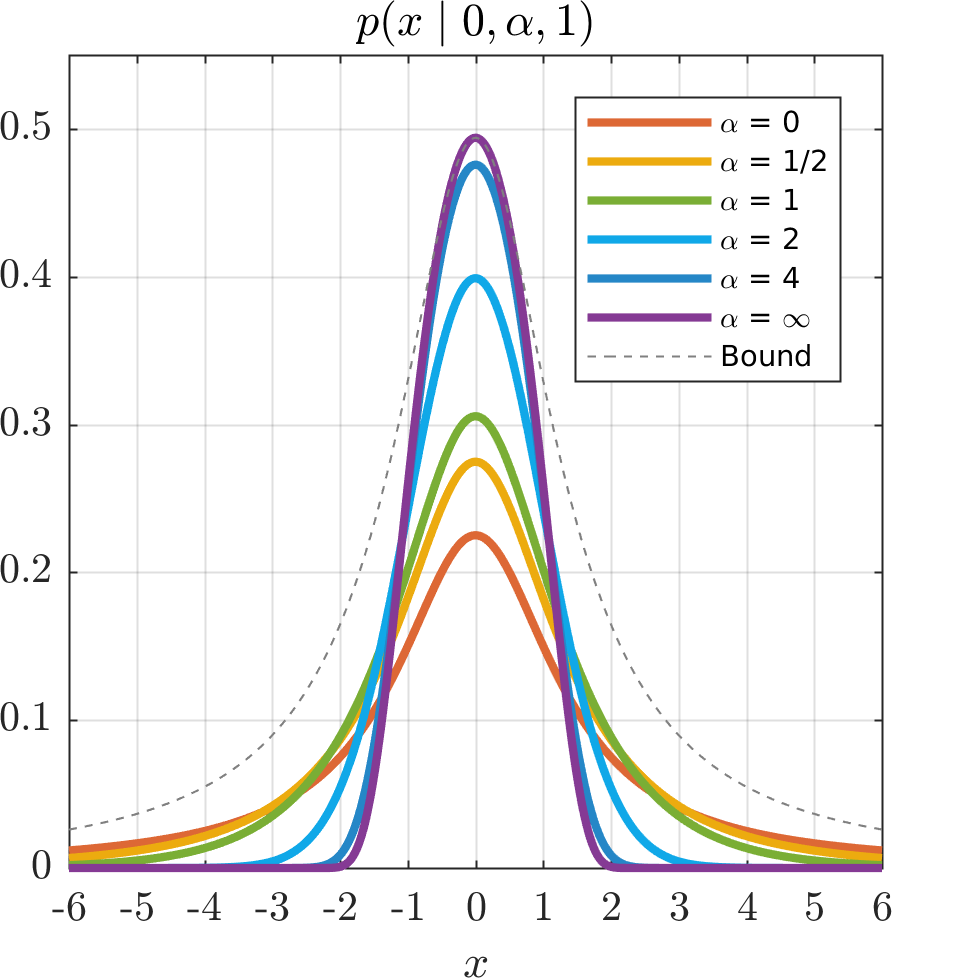}
  \end{tabular}
  \caption{
  The negative log-likelihoods (left) and
  probability densities (right) of the distribution corresponding to
  our loss function when it is defined ($\power \geq 0$).
  NLLs are simply losses (Fig.~\ref{fig:loss_and_grad}) shifted by a log partition
  function.
  Densities are bounded by a scaled Cauchy distribution.
  \label{fig:nll_prob}
  }
\end{figure}

Just as our loss function includes several common loss function as special
cases, our distribution includes several common distributions as special cases.
When $\power = 2$ our distribution becomes a normal (Gaussian) distribution, and
when $\power = 0$ our distribution becomes a Cauchy distribution.
These are also both special cases of Student's $t$-distribution ($\nu=\infty$ and $\nu = 1$, respectively),
though these are the only two points where these two families of distributions
intersect. Our distribution resembles the
generalized Gaussian distribution~\cite{nadarajah2005generalized, subbotin1923law},
except that it is ``smoothed'' so as to approach a Gaussian distribution near the origin regardless of the
shape parameter $\power$. The PDF and NLL of our distribution for different
values of $\power$ can be seen in Figure~\ref{fig:nll_prob}.

In later experiments we will use the NLL of our general distribution $-\log(p(\cdot | \alpha, c))$
as the loss for training our neural networks, not our general loss $\lossfun{\cdot, \power, c}$.
Critically, using the NLL allows us to treat $\power$ as a free parameter, thereby allowing optimization to
automatically determine the degree of robustness that should be imposed by the loss being used during training.
To understand why the NLL must be used for this, consider a training procedure
in which we simply minimize $\lossfun{\cdot, \power, c}$ with
respect to $\power$ and our model weights.
In this scenario, the monotonicity of our general loss with respect to $\power$ (Eq.~\ref{eq:monopower})
means that optimization can trivially minimize the cost of outliers by setting $\power$
to be as small as possible.
Now consider that same training procedure in which we minimize the NLL of our
distribution instead of our loss. As can be observed in Figure~\ref{fig:nll_prob},
reducing $\power$ will decrease the NLL of outliers but will \emph{increase} the NLL
of inliers. During training, optimization will have to choose between reducing $\power$,
thereby getting ``discount'' on large errors at the cost of paying a penalty for small errors,
or increasing  $\power$, thereby incurring a higher cost for outliers but a lower cost for inliers.
This tradeoff forces optimization to judiciously adapt the robustness of the
NLL being minimized.
As we will demonstrate later, allowing the NLL to adapt in this way can
increase performance on a variety of learning tasks, in addition to obviating
the need for manually tuning $\power$ as a fixed hyperparameter.

Sampling from our distribution is straightforward given
the observation that $-\log\left(\prob{x \;|\; 0, \power, 1}\right)$ is bounded
from below by $\rho(x, 0, 1) + \log(Z(\power))$ (shifted Cauchy loss).
See Figure~\ref{fig:nll_prob} for visualizations of this bound when $\power=\infty$,
which also bounds the NLL for all values of $\power$.
This lets us perform rejection sampling using a Cauchy as the proposal distribution.
Because our distribution is a location-scale family, we sample
from $\prob{x \;|\; 0, \power, 1}$ and then scale and shift that
sample by $c$ and $\mu$ respectively.
This sampling approach is efficient, with an acceptance rate between $\sim\!45\%$ ($\power=\infty$)
and $100\%$ ($\power = 0$).
Pseudocode for sampling is shown in Algorithm~\ref{alg:sample}.

\begin{algorithm}[b!]
\caption{Sampling from our general distribution
\label{alg:sample}}
{\bf Input:} Parameters for the distribution to sample $\{\mu, \power, c \}$
\\
{\bf Output:} A sample drawn from $\prob{x \;|\; \mu, \power, c}$.
\begin{algorithmic}[1]
\State \algorithmicwhile\ True:
\State \quad $x \sim \operatorname {Cauchy} (x_0 = 0, \gamma = \sqrt{2})$
\State \quad $u \sim \operatorname {Uniform}(0, 1)$
\State \quad \algorithmicif\ $u  < {\prob{x \;|\; 0, \power, 1} \over \exp \left(-\rho(x, 0, 1) - \log(Z(\power)) \right)}$:
\State \quad\quad \algorithmicreturn\ $cx + \mu$
\end{algorithmic}
\end{algorithm}

\section{Experiments}
\label{sec:experiments}

We will now demonstrate the utility of our loss function and distribution with four experiments.
None of these results are intended to represent the state-of-the-art for any particular task ---
our goal is to demonstrate the value of our loss and distribution as useful tools in isolation.
We will show that across a variety of tasks, just replacing the loss function of
an existing model with our general loss function can enable significant performance improvements.

In Sections~\ref{sec:vae} and \ref{sec:monodepth} we focus on learning
based vision tasks in which training involves minimizing the difference between images:
variational autoencoders for image synthesis and self-supervised monocular depth estimation.
We will generalize and improve models for both tasks  by using our general distribution
(either as a conditional distribution in a generative model or by using its NLL as an adaptive loss)
and allowing the distribution to \emph{automatically} determine its own degree of robustness.
Because robustness is automatic and requires no manually-tuned hyperparameters, we can even
allow for the robustness of our loss to be adapted individually for each dimension of our
output space --- we can have a different degree of robustness at each pixel in an image, for example.
As we will show, this approach is particularly effective when combined with image representations
such as wavelets, in which we expect to see non-Gaussian, heavy-tailed distributions.

In Sections~\ref{sec:registration} and \ref{sec:rcc}
we will build upon existing algorithms for two classic vision tasks (registration and clustering)
that both work by minimizing a robust loss that is subsumed by our general loss.
We will then replace each algorithm's fixed robust loss with our loss, thereby introducing a
continuous tunable robustness parameter $\power$.
This generalization allows us to introduce new models in which $\power$ is manually tuned or annealed,
thereby improving performance.
These results demonstrate the value of our loss function when
designing classic vision algorithms, by allowing model robustness to be introduced
into the algorithm design space as a continuous hyperparameter.

\subsection{Variational Autoencoders}
\label{sec:vae}

Variational autoencoders \cite{KingmaW13, rezende14} are a landmark technique for training
autoencoders as generative models, which can then be used to draw random samples that resemble
training data. We will demonstrate that our general
distribution can be used to improve the log-likelihood performance of VAEs for image synthesis on
the CelebA dataset~\cite{Liu2015}.
A common design decision for VAEs is to model images using an independent normal
distribution on a vector of RGB pixel values \cite{KingmaW13}, and we use this design as our baseline model.
Recent work has improved upon this model by using deep, learned, and adversarial
loss functions \cite{DosovitskiyNIPS2016, NIPS2014, Larsen2016}.
Though it's possible that our general loss or distribution can add value in these circumstances,
to more precisely isolate our contribution we will
explore the hypothesis that the baseline model of normal distributions placed on a per-pixel image representation
can be improved significantly with the small change of just modeling a linear transformation of a VAE's output
with our general distribution.
Again, our goal is not to advance the state of the art for any particular image
synthesis task, but is instead to explore the value of our distribution in an
experimentally controlled setting.

In our baseline model we give each pixel's normal distribution a variable scale parameter $\sigma^{(i)}$ that will
be optimized over during training, thereby allowing the VAE to adjust
the scale of its distribution for each output dimension.
We can straightforwardly replace this per-pixel normal distribution with a per-pixel
general distribution, in which each output dimension is given
a distinct shape parameter $\power^{(i)}$ in addition to its scale parameter $c^{(i)}$ (\ie, $\sigma^{(i)}$).
By letting the $\power^{(i)}$ parameters be free variables alongside the scale parameters,
training is able to adaptively select both the scale and robustness of the VAE's posterior
distribution over pixel values.
We restrict all $\power^{(i)}$ to be in $(0, 3)$, which allows our distribution to generalize
Cauchy ($\power=0$) and Normal ($\power = 2$) distributions and anything in between, as well as
more platykurtic distributions ($\power > 2$) which helps for this task.
We limit $\power$ to be less than $3$ because of the increased risk of numerical instability
during training as $\power$ increases.
We also compare against a Cauchy distribution as an example of a fixed heavy-tailed distribution, and against
Student's t-distribution as an example of a distribution that can adjust its own
robustness similarly to ours.

\newcommand{\amax}{\power_{\mathrm{max}}}
\newcommand{\amin}{\power_{\mathrm{min}}}
\newcommand{\cmin}{c_{\mathrm{min}}}

Regarding implementation, for each output dimension $i$ we construct
unconstrained TensorFlow variables $\{ \power_{\ell}^{(i)} \}$ and $\{ c_{\ell}^{(i)} \}$ and define
\begin{align}
\power^{(i)} &= (\amax - \amin) \operatorname{sigmoid}\left(\power_{\ell}^{(i)}\right) + \amin \\
c^{(i)} &= \operatorname{softplus}\left(c_{\ell}^{(i)}\right) + \cmin \\
& \amin = 0, \,\, \amax = 3, \,\, \cmin = 10^{-8}
\end{align}
The $\cmin$ offset avoids degenerate optima where
likelihood is maximized by having $c^{(i)}$ approach $0$, while $\amin$ and $\amax$
determine the range of values that $\power^{(i)}$ can take.
Variables are initialized such that initially all $\power^{(i)}=1$ and $c^{(i)}=0.01$,
and are optimized simultaneously with the autoencoder's weights using the same
Adam \cite{KingmaB14} optimizer instance.

Though modeling images using independent distributions on pixel intensities
is a popular choice due to its simplicity, classic work in natural image statistics suggest that
images are better modeled with heavy-tailed distributions on wavelet-like image decompositions \cite{Field87, Mallat1989}.
We therefore train additional models in which
our decoded RGB per-pixel images are linearly transformed into spaces that
better model natural images before computing the NLL of our distribution.
For this we use the DCT \cite{ahmed1974discrete}
and the CDF 9/7 wavelet decomposition \cite{cohen1992biorthogonal}, both with a YUV colorspace.
These representations resemble the JPEG and JPEG 2000 compression standards, respectively.

\definecolor{Yellow}{rgb}{1,1, 0.6}
\definecolor{Orange}{rgb}{1,0.8, 0.6}
\definecolor{Red}{rgb}{1, 0.6, 0.6}

\newcommand{\boxnum}[1]{\makebox[\widthof{37{,}000}][r]{#1}}

\begin{table}
\centering
\large
\resizebox{\linewidth}{!}{%
\begin{tabular}{l |c @{\hspace{0.002in}} c  c  c  c}
 & & Normal & Cauchy & t-dist. & Ours  \\ \hline
Pixels + RGB &&
\boxnum{8{,}662} &
\boxnum{9{,}602} &
{\cellcolor{Yellow} \boxnum{10{,}177}} &
{\cellcolor{Orange} \boxnum{10{,}240}} \\
DCT + YUV    &&
\boxnum{31{,}837} &
\boxnum{31{,}295} &
{\cellcolor{Yellow} \boxnum{32{,}804}} &
{\cellcolor{Orange} \boxnum{32{,}806}} \\
Wavelets + YUV &&
\boxnum{31{,}505} &
\boxnum{35{,}779} &
{\cellcolor{Orange} \boxnum{36{,}373}} &
{\cellcolor{Yellow} \boxnum{36{,}316}}
\end{tabular}
}
\vspace{0.02in}
\caption{Validation set ELBOs (higher is better) for our variational autoencoders.
Models using our general distribution better maximize the likelihood of unseen data
than those using normal or Cauchy distributions (both special cases of our model)
for all three image representations,
and perform similarly to Student's t-distribution (a different generalization of normal and Cauchy distributions).
The best and second best performing techniques for each representation are colored orange and yellow respectively.
}
\label{table:elbos}
\end{table}

\newcommand{\vaewidth}{0.238\linewidth}
\begin{figure}  \centering

{
\setlength{\tabcolsep}{0em} 
\renewcommand{\arraystretch}{0}

\begin{tabular}{@{}l@{\hspace{0.04in}}c@{\hspace{0in}}c@{\hspace{0in}}c@{\hspace{0in}}c@{}}
    &
    Normal
    &
    Cauchy
    &
    t-distribution
    &
    Ours
    \\
    \rotatebox[origin=l]{90}{\,\,\,\,\,\,Pixels + RGB}
    &
    \includegraphics[width=\vaewidth]{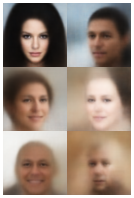}
    &
    \includegraphics[width=\vaewidth]{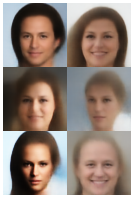}
    &
    \includegraphics[width=\vaewidth]{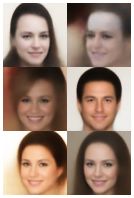}
    &
    \includegraphics[width=\vaewidth]{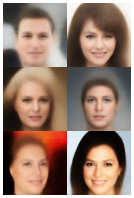}
    \\
    \rotatebox[origin=l]{90}{\,\,\,\,\,\,\,DCT + YUV}
    &
    \includegraphics[width=\vaewidth]{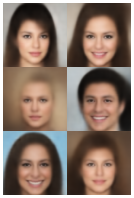}
    &
    \includegraphics[width=\vaewidth]{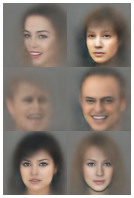}
    &
    \includegraphics[width=\vaewidth]{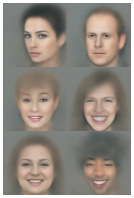}
    &
    \includegraphics[width=\vaewidth]{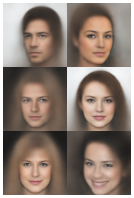}
    \\
    \rotatebox[origin=l]{90}{\,\, Wavelets + YUV}
    &
    \includegraphics[width=\vaewidth]{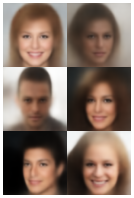}
    &
    \includegraphics[width=\vaewidth]{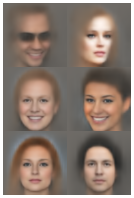}
    &
    \includegraphics[width=\vaewidth]{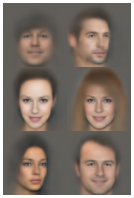}
    &
    \includegraphics[width=\vaewidth]{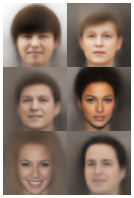}
  \end{tabular}
  }
  \caption{Random samples
   from our variational autoencoders.
    We use either normal, Cauchy, Student's t, or our general distributions (columns) to model the
    coefficients of three different image representations (rows).
    Because our distribution can adaptively interpolate between
    Cauchy-like or normal-like behavior for each coefficient individually,
    using it results in sharper and higher-quality samples (particularly
    when using DCT or wavelet representations) and does a better job of capturing
    low-frequency image content than Student's t-distribution.
  \label{fig:vae_samples}
  }
\end{figure}

Our results can be seen in Table~\ref{table:elbos}, where we
report the validation set evidence lower bound (ELBO) for all combinations of our four
distributions and three image representations, and in Figure~\ref{fig:vae_samples},
where we visualize samples from these models. We see that
our general distribution performs similarly to a Student's t-distribution,
with both producing higher ELBOs than any fixed distribution across all representations.
These two adaptive distributions appear to have complementary strengths: ours can be more platykurtic
($\alpha>2$) while a t-distribution can be more leptokurtic ($\nu < 1$),
which may explain why neither model consistently outperforms the other across
representations.
Note that the t-distribution's NLL does not generalize the Charbonnier, L1, Geman-McClure,
or Welsch losses, so unlike ours it will not generalize the losses used in the other tasks we will address.
For all representations, VAEs trained with our general distribution
produce sharper and more detailed samples than those trained with normal distributions.
Models trained with Cauchy and t-distributions preserve high-frequency detail and work well
on pixel representations, but systematically fail to synthesize low-frequency image content
when given non-pixel representations,
as evidenced by the gray backgrounds of those samples.
Comparing performance across image representations shows that the ``Wavelets + YUV''
representation best maximizes validation set ELBO --- though if we were to limit our
model to only normal distributions the ``DCT + YUV'' model would appear superior, suggesting
that there is value in reasoning jointly about distributions and image representations.
After training we see
shape parameters $\{ \power^{(i)} \}$ that span $(0, 2.5)$, suggesting that an adaptive mixture of
normal-like and Cauchy-like distributions is useful in modeling natural images,
as has been observed previously~\cite{PortillaSWS03}.
Note that this adaptive robustness is just a consequence of allowing $\{\power_{\ell}^{(i)}\}$
to be free variables during training, and requires no manual parameter tuning.
See Appendix~\ref{app:vae} for more samples and reconstructions from these models, and a
review of our experimental procedure.

\subsection{Unsupervised Monocular Depth Estimation}
\label{sec:monodepth}

Due to the difficulty of acquiring ground-truth direct depth observations,
there has been recent interest in ``unsupervised'' monocular depth estimation,
in which stereo pairs and geometric constraints are used to directly train
a neural network \cite{DeepStereo, garg2016unsupervised, monodepth17, zhou2017}.
We use \cite{zhou2017} as a representative model from this literature,
which is notable for its estimation of depth \emph{and} camera pose.
This model is trained by minimizing the differences between two images in a stereo pair,
where one image has been warped to match the other according to the depth and pose
predictions of a neural network.
In \cite{zhou2017} that difference between images is defined as the absolute difference between RGB values.
We will replace that loss with different varieties of our general loss,
and demonstrate that using annealed or adaptive forms of our loss can improve performance.

The absolute loss in \cite{zhou2017} is equivalent to maximizing the
likelihood of a Laplacian distribution with a fixed scale on RGB pixel values.
We replace that fixed Laplacian distribution with our general distribution,
keeping our scale fixed but allowing the shape parameter $\power$ to vary.
Following our observation from Section~\ref{sec:vae} that
YUV wavelet representations work well when modeling images with our loss function, we
impose our loss on a YUV wavelet decomposition instead of the RGB pixel representation of \cite{zhou2017}.
The only changes we made to the code from \cite{zhou2017} were to replace its
loss function with our own and to remove the model components that stopped yielding any
improvement after the loss function was replaced (see Appendix~\ref{app:monodepth} for details).
All training and evaluation was performed on the KITTI dataset \cite{kitti}
using the same training/test split as \cite{zhou2017}.

\begin{table}
\centering
\Huge
\resizebox{\linewidth}{!}{%
\begin{tabular}{@{}l|c||cccc|ccc@{}}
& \multicolumn{5}{c|}{lower is better} & \multicolumn{3}{c}{higher is better} \\
& Avg & AbsRel & SqRel & RMS & logRMS & $<\!1.25$ & $<\!1.25^2$ & $<\!1.25^3$  \\ \hline
    Baseline \cite{zhou2017} as reported & 0.407 & 0.221 & 2.226 & 7.527 & 0.294 & 0.676 & 0.885 & 0.954 \\
     Baseline \cite{zhou2017} reproduced & 0.398 & 0.208 & 2.773 & 7.085 & 0.286 & 0.726 & 0.895 & 0.953 \\
                  Ours, fixed $\power=1$ & 0.356 & 0.194 & 2.138 \cellcolor{Yellow} & 6.743 & 0.268 & 0.738 & 0.906 & 0.960 \\
                  Ours, fixed $\power=0$ & 0.350 & 0.187 \cellcolor{Yellow} & 2.407 & 6.649 \cellcolor{Yellow} & 0.261 \cellcolor{Yellow} & 0.766 \cellcolor{Orange} & 0.911 \cellcolor{Yellow} & 0.960 \\
                  Ours, fixed $\power=2$ & 0.349 \cellcolor{Yellow} & 0.190 & 1.922 \cellcolor{Red} & 6.648 \cellcolor{Orange} & 0.267 & 0.737 & 0.904 & 0.961 \cellcolor{Yellow} \\
Ours, annealing $\power=2\!\rightarrow\!0$ & 0.341 \cellcolor{Orange} & 0.184 \cellcolor{Orange} & 2.063 \cellcolor{Orange} & 6.697 & 0.260 \cellcolor{Orange} & 0.756 \cellcolor{Yellow} & 0.911 \cellcolor{Orange} & 0.963 \cellcolor{Orange} \\
      Ours, adaptive $\power \in (0, 2)$ & 0.332 \cellcolor{Red} & 0.181 \cellcolor{Red} & 2.144 & 6.454 \cellcolor{Red} & 0.254 \cellcolor{Red} & 0.766 \cellcolor{Red} & 0.916 \cellcolor{Red} & 0.965 \cellcolor{Red} \\
\end{tabular}
}
\vspace{0.02in}
\caption{
Results on unsupervised monocular depth estimation using the KITTI dataset \cite{kitti},
building upon the model from \cite{zhou2017} (``Baseline'').
By replacing the per-pixel loss used by \cite{zhou2017} with several variants of our own per-wavelet general
loss function in which our loss's shape parameters are fixed, annealed, or adaptive, we
see a significant performance improvement.
The top three techniques are colored red, orange, and yellow for each metric.
\label{table:sfmlearner}
}
\end{table}

\newcommand{\sfmwidth}{0.93\linewidth}
\newcommand{\nametag}{153}
\begin{figure}[t]
  \centering
  \begin{tabular}{@{}l@{\hspace{0.04in}}c@{}}
    \rotatebox[origin=l]{90}{\,\,\,\,\,\,Input}
    &
    \includegraphics[width=\sfmwidth]{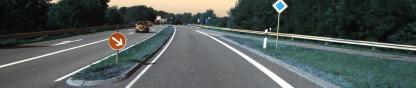}
    \\
    \rotatebox[origin=l]{90}{\,Baseline}
    &
    \includegraphics[width=\sfmwidth]{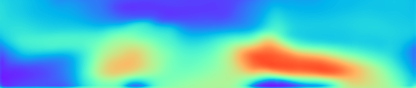}
    \\
    \rotatebox[origin=l]{90}{\,\,\,\,\,\,Ours}
    &
    \includegraphics[width=\sfmwidth]{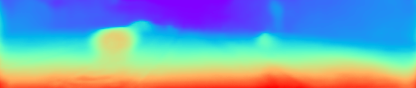}
    \\
    \rotatebox[origin=l]{90}{\,\,\,\,Truth}
    &
    \includegraphics[width=\sfmwidth]{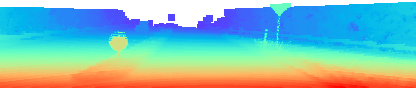}
    \\
  \end{tabular}
  \caption{
  Monocular depth estimation results on the KITTI benchmark using the ``Baseline'' network of \cite{zhou2017}.
  Replacing only the network's loss function with our ``adaptive'' loss over wavelet coefficients
  results in significantly improved depth estimates.
  \label{fig:sfm_results}
  }
\end{figure}

Results can be seen in Table~\ref{table:sfmlearner}.
We present the error and accuracy metrics used in \cite{zhou2017} and our own ``average'' error measure:
the geometric mean of the four errors and one minus the three accuracies.
The ``Baseline`` models use the loss function of \cite{zhou2017}, and we present
both the numbers in \cite{zhou2017} (``as reported'') and our own numbers
from running the code from \cite{zhou2017} ourselves (``reproduced'').
The ``Ours'' entries all use our general loss imposed on wavelet
coefficients, but for each entry we use a different strategy
for setting the shape parameter or parameters. We keep our loss's scale $c$
fixed to $0.01$, thereby matching the fixed scale assumption of the baseline model and
roughly matching the shape of its L1 loss (Eq.~\ref{eq:L1loss}).
To avoid exploding gradients we multiply the loss being minimized by $c$, thereby bounding
gradient magnitudes by residual magnitudes (Eq.~\ref{eq:gradbound}).
For the ``fixed'' models we use a constant value for $\power$ for all wavelet
coefficients, and observe that though performance is improved relative to the baseline,
no single value of $\power$ is optimal. The $\power=1$ entry is simply
a smoothed version of the L1 loss used by the baseline model,
suggesting that just using a wavelet representation improves performance.
In the ``annealing $\power=2\!\rightarrow\!0$'' model we linearly interpolate $\power$
from $2$ (L2) to $0$ (Cauchy) as a function of training iteration,
which outperforms all ``fixed'' models.
In the ``adaptive $\power \in (0, 2)$'' model we assign each wavelet
coefficient its own shape parameter as a free variable and we allow those
variables to be optimized alongside our network weights during training
as was done in Section~\ref{sec:vae},
but with $\amin=0$ and $\amax=2$.
This ``adaptive'' strategy outperforms the ``annealing'' and all ``fixed'' strategies,
thereby demonstrating the value of allowing the model to adaptively determine the
robustness of its loss during training.
Note that though the ``fixed'' and ``annealed'' strategies only require our
general loss, the ``adaptive'' strategy requires that we use the NLL of our general
distribution as our loss --- otherwise training
would simply drive $\power$ to be as small as possible due to the monotonicity
of our loss with respect to $\power$, causing performance to degrade to the ``fixed $\power=0$'' model.
Comparing the ``adaptive'' model's performance to that of the ``fixed'' models
suggests that, as in Section~\ref{sec:vae},
no single setting of $\power$ is optimal for all wavelet coefficients.
Overall, we see that just replacing the loss function of \cite{zhou2017} with our
adaptive loss on wavelet coefficients reduces average error by $\sim\!17\%$.

In Figure~\ref{fig:sfm_results} we compare our
``adaptive'' model's output to the baseline model and the ground-truth depth,
and demonstrate a substantial qualitative improvement.
See Appendix~\ref{app:monodepth} for many more results, and for visualizations of
the per-coefficient robustness selected by our model.

\subsection{Fast Global Registration}
\label{sec:registration}

Robustness is often a core component of geometric registration~\cite{RobustBA}.
The Fast Global Registration (FGR) algorithm of \cite{Zhou16} finds the rigid
transformation $\mathbf{T}$ that aligns point sets $\{ \mathbf{p} \}$ and $\{ \mathbf{q} \}$
by minimizing the following loss:
\begin{equation}
\sum_{ (\mathbf{p}, \mathbf{q})} \rho_{gm} \left( \norm{\mathbf{p} - \mathbf{T}\mathbf{q}}, c \right)
\end{equation}
where $\rho_{gm}(\cdot)$ is Geman-McClure loss.
By using the Black and Rangarajan duality between robust estimation and line processes \cite{BlackR96}
FGR is capable of producing high-quality registrations at high speeds.
Because Geman-McClure loss is a special case of our loss, and because we
can formulate our loss as an outlier process (see Appendix~\ref{app:alternative}), we can generalize FGR to an arbitrary shape parameter $\power$
by replacing $\rho_{gm}(\cdot, c)$ with our $\rho(\cdot, \power, c)$ (where setting $\power=-2$ reproduces FGR).

\begin{table}
\resizebox{0.98\linewidth}{!}{%
\begin{tabular}{l|ccc|ccc}
& \multicolumn{3}{c|}{Mean RMSE $\times 100$} & \multicolumn{3}{c}{Max RMSE $\times 100$} \\
\multicolumn{1}{r|}{$\sigma = $} & $0$ & $0.0025$ & $0.005$ & $0$ & $0.0025$ & $0.005$ \\
\hline \hline
   FGR \cite{Zhou16} & 0.373 & 0.518 & 0.821 & 0.591 & 1.040 & 1.767\\
   \hline
 shape-annealed gFGR & 0.374 & 0.510 & {\bf 0.802} & 0.590 & 0.997 & 1.670\\
               gFGR* & {\bf 0.370} & {\bf 0.509} & 0.806 & {\bf 0.545} & {\bf 0.961} & {\bf 1.669}\\
\end{tabular}
}
\vspace{0.02in}
\caption{
Results on the registration task of \cite{Zhou16}, in which we compare their ``FGR'' algorithm to two versions of our ``gFGR'' generalization.
}
\label{table:registration_results}
\end{table}

\begin{figure}[t!]
  \includegraphics[width=1.6in]{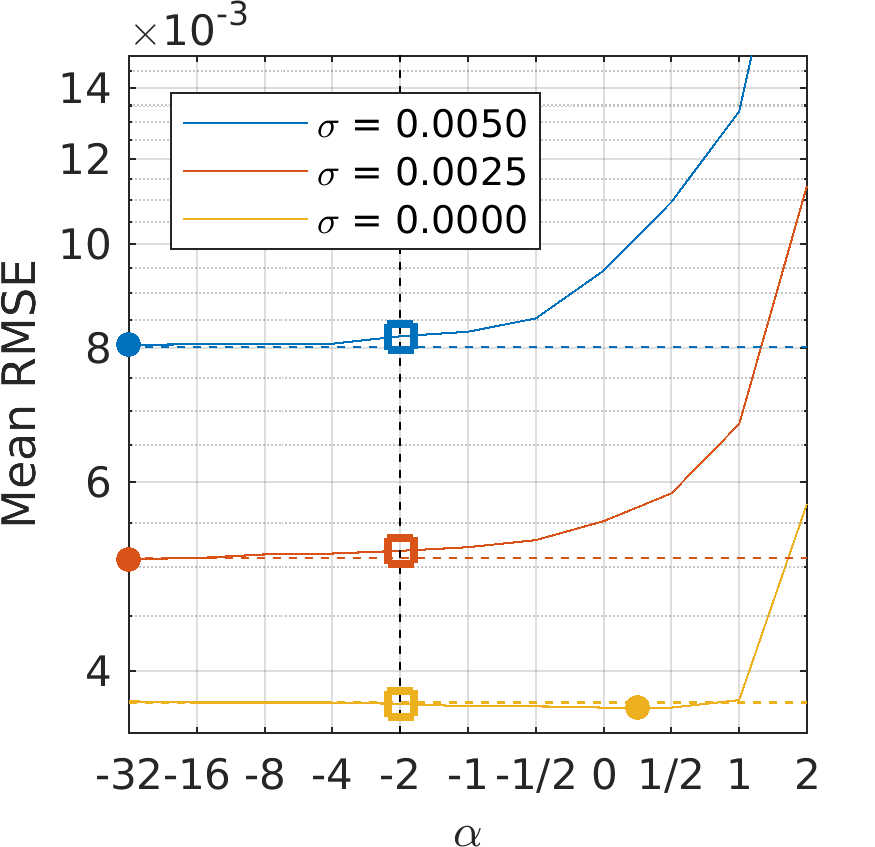}
  \includegraphics[width=1.6in]{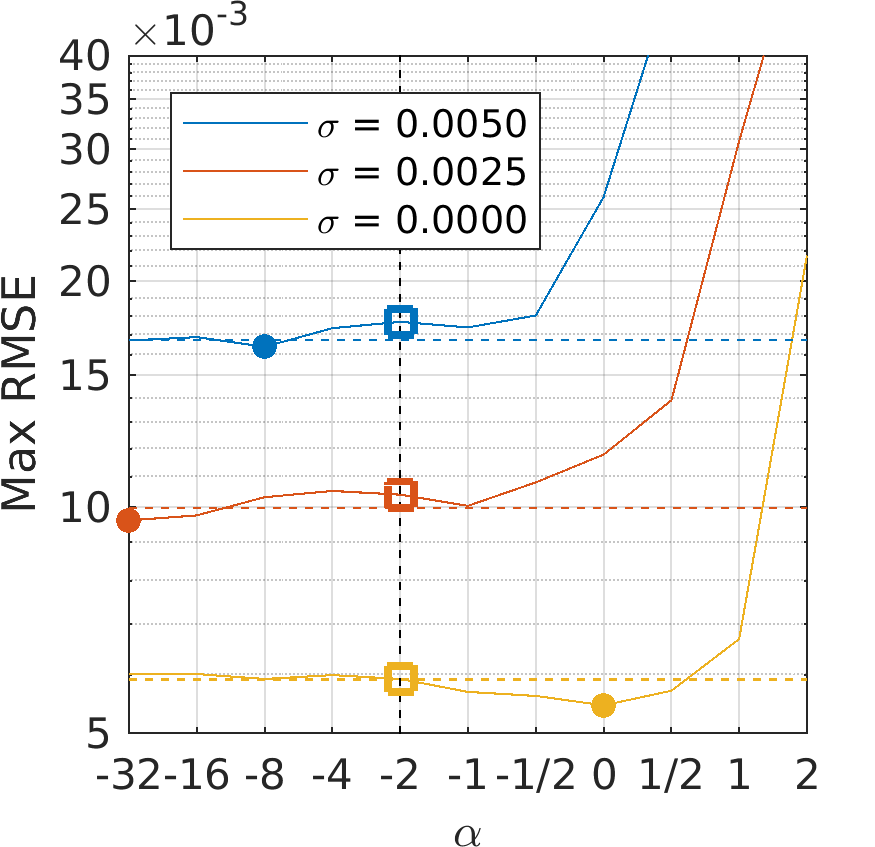}
\caption{
Performance (lower is better) of our gFGR algorithm on the task of \cite{Zhou16}
as we vary our shape parameter $\power$, with the lowest-error point indicated by a circle.
FGR (equivalent to gFGR with $\power=-2$) is shown as a dashed line and a square, and
shape-annealed gFGR for each noise level is shown as a dotted line.
}
\label{figure:registration_results}
\end{figure}

This generalized FGR (gFGR) enables algorithmic improvements.
FGR iteratively solves a linear system while annealing its scale parameter $c$, which has the effect of gradually introducing nonconvexity.
gFGR enables an alternative strategy in which we directly manipulate convexity by annealing $\power$ instead of $c$.
This ``shape-annealed gFGR'' follows the same procedure as \cite{Zhou16}: $64$ iterations in which a parameter is annealed every $4$ iterations.
Instead of annealing $c$, we set it to its terminal value and instead anneal $\power$ over the following values:
\begin{equation}
2, 1, \nicefrac{1}{2}, \nicefrac{1}{4}, 0, -\nicefrac{1}{4}, -\nicefrac{1}{2}, -1, -2, -4, -8, -16, -32 \nonumber
\end{equation}
Table~\ref{table:registration_results} shows results for the 3D point cloud registration task of \cite{Zhou16} (Table 1 in that paper),
which shows that annealing shape produces moderately improved performance over FGR for high-noise inputs,
and behaves equivalently in low-noise inputs.
This suggests that performing graduated non-convexity by directly adjusting a shape parameter that controls non-convexity
--- a procedure that is enabled by our general loss --
is preferable to indirectly controlling non-convexity by annealing a scale parameter.

Another generalization is to continue using the $c$-annealing strategy of \cite{Zhou16},
but treat $\power$ as a hyperparameter and tune it independently for each noise level in this task.
In Figure~\ref{figure:registration_results} we set $\power$ to a wide
range of values and report errors for each setting, using the same evaluation of \cite{Zhou16}.
We see that for high-noise inputs more negative values of $\power$ are preferable,
but for low-noise inputs values closer to $0$ are optimal.
We report the lowest-error entry for each noise level as ``gFGR*'' in Table~\ref{table:registration_results}
where we see a significant reduction in error, thereby demonstrating the improvement
that can be achieved from treating robustness as a hyperparameter.

\subsection{Robust Continuous Clustering}
\label{sec:rcc}

\begin{figure}[t!]
  \centering
  \includegraphics[width=2.5in]{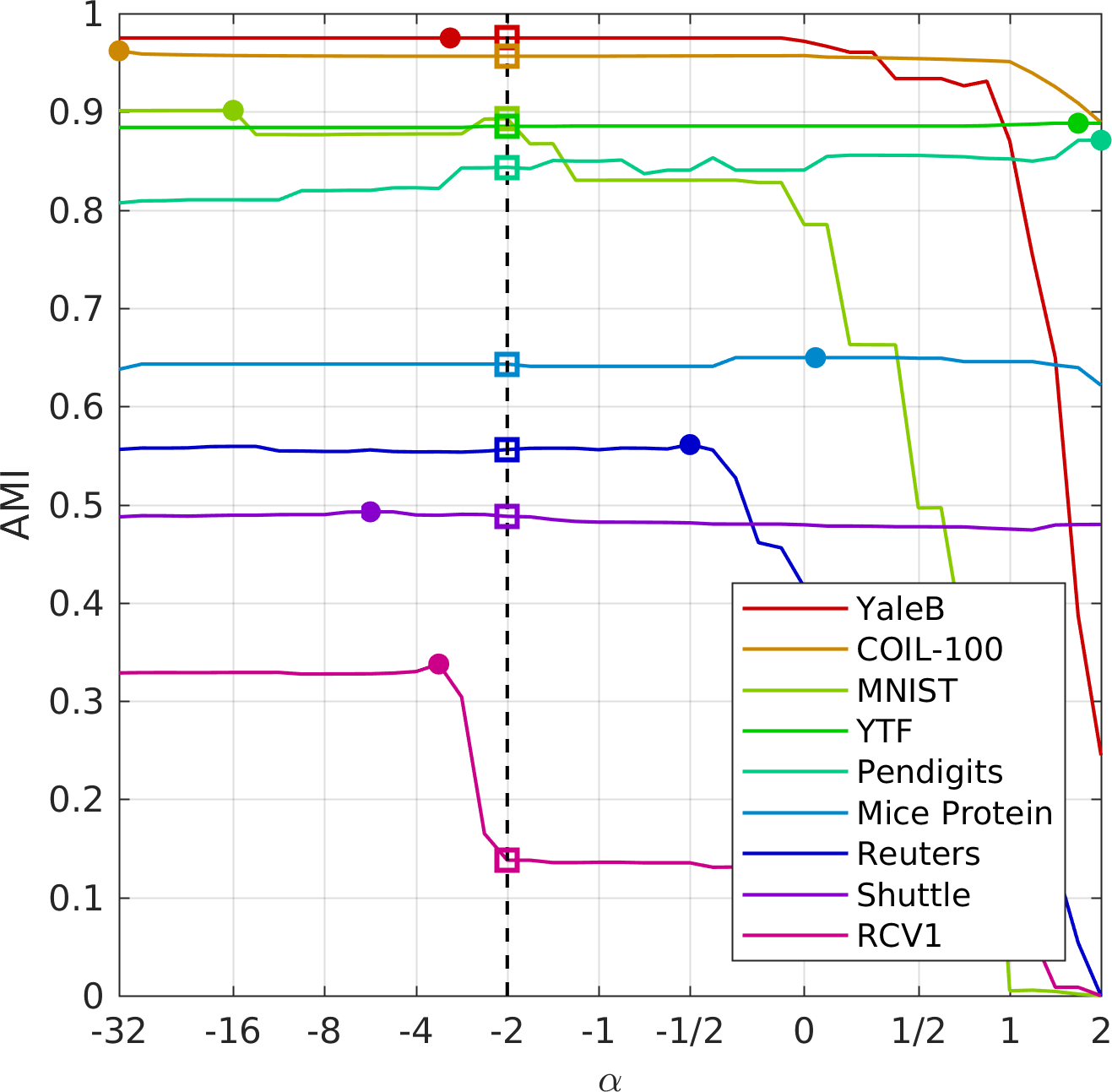}
\caption{
Performance (higher is better) of our gRCC algorithm on the clustering task of \cite{shah2017robust},
for different values of our shape parameter $\power$, with the highest-accuracy point indicated by a dot.
Because the baseline RCC algorithm is equivalent to gRCC with $\power=-2$,
we highlight that $\power$ value with a dashed line and a square.
}
\label{figure:rcc_results}
\end{figure}

In \cite{shah2017robust} robust losses are used for unsupervised clustering, by minimizing:
\begin{equation}
\sum_{i} \norm{ \mathbf{x}_i - \mathbf{u}_i}_2^2 + \lambda \sum_{ (p, q) \in \mathcal{E} } w_{p,q} \rho_{gm}\left( \norm{ \mathbf{u}_p - \mathbf{u}_q}_2 \right)
\end{equation}
where $\{ \mathbf{x}_i \}$ is a set of input datapoints,
$\{ \mathbf{u}_i \}$ is a set of ``representatives'' (cluster centers),
and $\mathcal{E}$ is a mutual k-nearest neighbors (m-kNN) graph.
As in Section~\ref{sec:registration}, $\rho_{gm}(\cdot)$ is Geman-McClure loss,
which means that our loss can be used to generalize this algorithm.
Using the RCC code provided by the authors (and keeping all hyperparameters fixed to their default values)
we replace Geman-McClure loss with our general loss and then sweep over values of $\power$.
In Figure~\ref{figure:rcc_results} we show the adjusted mutual information
(AMI, the metric used by \cite{shah2017robust}) of the resulting clustering
for each value of $\power$ on the datasets used in \cite{shah2017robust}, and in
Table~\ref{table:rcc_results} we report the AMI for the best-performing
value of $\power$ for each dataset as ``gRCC*''.
On some datasets performance is insensitive to $\power$, but on others
adjusting $\power$ improves performance by as much as $32\%$.
This improvement demonstrates the gains that can be achieved by introducing
robustness as a hyperparameter and tuning it accordingly.

\begin{table}
\resizebox{0.98\linewidth}{!}{%
\begin{tabular}{l||cccccc|cc}
     Dataset & \rotatebox[origin=l]{90}{AC-W} & \rotatebox[origin=l]{90}{N-Cuts~\cite{ShiM00}} & \rotatebox[origin=l]{90}{LDMGI~\cite{YangXNYZ10}} & \rotatebox[origin=l]{90}{PIC~\cite{ZHANG20133056}} & \rotatebox[origin=l]{90}{RCC-DR~\cite{shah2017robust}} & \rotatebox[origin=l]{90}{RCC~\cite{shah2017robust}} & \rotatebox[origin=l]{90}{gRCC*} & \rotatebox[origin=l]{90}{Rel. Impr.} \\ \hline
       YaleB & $0.767$ & $0.928$ & $0.945$ & $0.941$ & $0.974$ & $0.975$ & $\mathbf{0.975}$ & \multicolumn{1}{r}{$0.4\%$} \\
    COIL-100 & $0.853$ & $0.871$ & $0.888$ & $\mathbf{0.965}$ & $0.957$ & $0.957$ & $0.962$ & \multicolumn{1}{r}{$11.6\%$} \\
       MNIST & $0.679$ & - & $0.761$ & - & $0.828$ & $0.893$ & $\mathbf{0.901}$ & \multicolumn{1}{r}{$7.9\%$} \\
         YTF & $0.801$ & $0.752$ & $0.518$ & $0.676$ & $0.874$ & $0.836$ & $\mathbf{0.888}$ & \multicolumn{1}{r}{$31.9\%$} \\
   Pendigits & $0.728$ & $0.813$ & $0.775$ & $0.467$ & $0.854$ & $0.848$ & $\mathbf{0.871}$ & \multicolumn{1}{r}{$15.1\%$} \\
Mice Protein & $0.525$ & $0.536$ & $0.527$ & $0.394$ & $0.638$ & $0.649$ & $\mathbf{0.650}$ & \multicolumn{1}{r}{$0.2\%$} \\
     Reuters & $0.471$ & $0.545$ & $0.523$ & $0.057$ & $0.553$ & $0.556$ & $\mathbf{0.561}$ & \multicolumn{1}{r}{$1.1\%$} \\
     Shuttle & $0.291$ & $0.000$ & $\mathbf{0.591}$ & - & $0.513$ & $0.488$ & $0.493$ & \multicolumn{1}{r}{$0.9\%$} \\
        RCV1 & $0.364$ & $0.140$ & $0.382$ & $0.015$ & $\mathbf{0.442}$ & $0.138$ & $0.338$ & \multicolumn{1}{r}{$23.2\%$} \\
\end{tabular}
}
\vspace{0.02in}
\caption{
Results on the clustering task of \cite{shah2017robust}
where we compare their ``RCC'' algorithm to our ``gRCC*'' generalization in terms of
AMI on several datasets. We also report the AMI increase of ``gRCC*'' with respect to ``RCC''.
Baselines are taken from \cite{shah2017robust}.
}
\label{table:rcc_results}
\end{table}

\section{Conclusion}
\label{sec:conclusion}

We have presented a two-parameter loss function that generalizes many existing one-parameter robust loss functions:
the Cauchy/Lorentzian, Geman-McClure, Welsch/Leclerc, generalized Charbonnier,
Charbonnier/pseudo-Huber/L1-L2, and L2 loss functions.
By reducing a family of discrete single-parameter losses to a
single function with two continuous parameters, our loss enables
the convenient exploration and comparison of different robust penalties.
This allows us to generalize and improve algorithms designed around the minimization of some
fixed robust loss function, which we have demonstrated for registration and clustering.
When used as a negative log-likelihood, this loss gives a general probability
distribution that includes normal and Cauchy distributions as special cases.
This distribution lets us train neural networks in which the loss has an adaptive degree
of robustness for each output dimension, which allows training
to automatically determine how much robustness should be imposed by the loss
without any manual parameter tuning.
When this adaptive loss is paired with image representations in which variable
degrees of heavy-tailed behavior occurs, such as wavelets, this adaptive
training approach allows us to improve the performance
of variational autoencoders for image synthesis and of
neural networks for unsupervised monocular depth estimation.

\paragraph{Acknowledgements:} Thanks to Rob Anderson, Jesse Engel, David Gallup, Ross Girshick, Jaesik Park, Ben Poole, Vivek Rathod, and Tinghui Zhou.
\clearpage

{\small
\bibliographystyle{ieee}
\bibliography{robust}
}

\clearpage

\begin{appendices}

\section{Alternative Forms}
\label{app:alternative}

The registration and clustering experiments in the paper require that we
formulate our loss as an outlier process.
Using the equivalence between robust loss minimization and outlier processes
established by Black and Rangarajan \cite{BlackR96}, we can derive our loss's $\Psi$-function:
\begin{align}
& \resizebox{\linewidth}{!}{$%
\Psi(z, \power) = \begin{cases}
-\log(z) + z - 1 & \text{if } \power = 0 \\
z \log(z) - z + 1  & \text{if } \power = -\infty \\
{\abs{\power - 2} \over \power} \left( \left( 1 - {\power \over 2} \right) z^{\power \over (\power-2)} + {\power z \over 2} - 1 \right)  & \text{if } \power < 2
\end{cases}
$} \nonumber \\ %
& \quad\quad \rho(x, \power, c) = \min_{0 \leq z \leq 1} \left( {1 \over 2} \left( \sfrac{x}{c} \right)^2 z + \Psi(z, \power) \right)
\end{align}
$\Psi(z, \power)$ is not defined when $\power \geq 2$ because for those values the loss is no longer robust,
and so is not well described as a process that rejects outliers.

\begin{figure}[t]
\centering
\begin{tabular}{@{}c@{}c@{}}
  \includegraphics[width=0.5\linewidth]{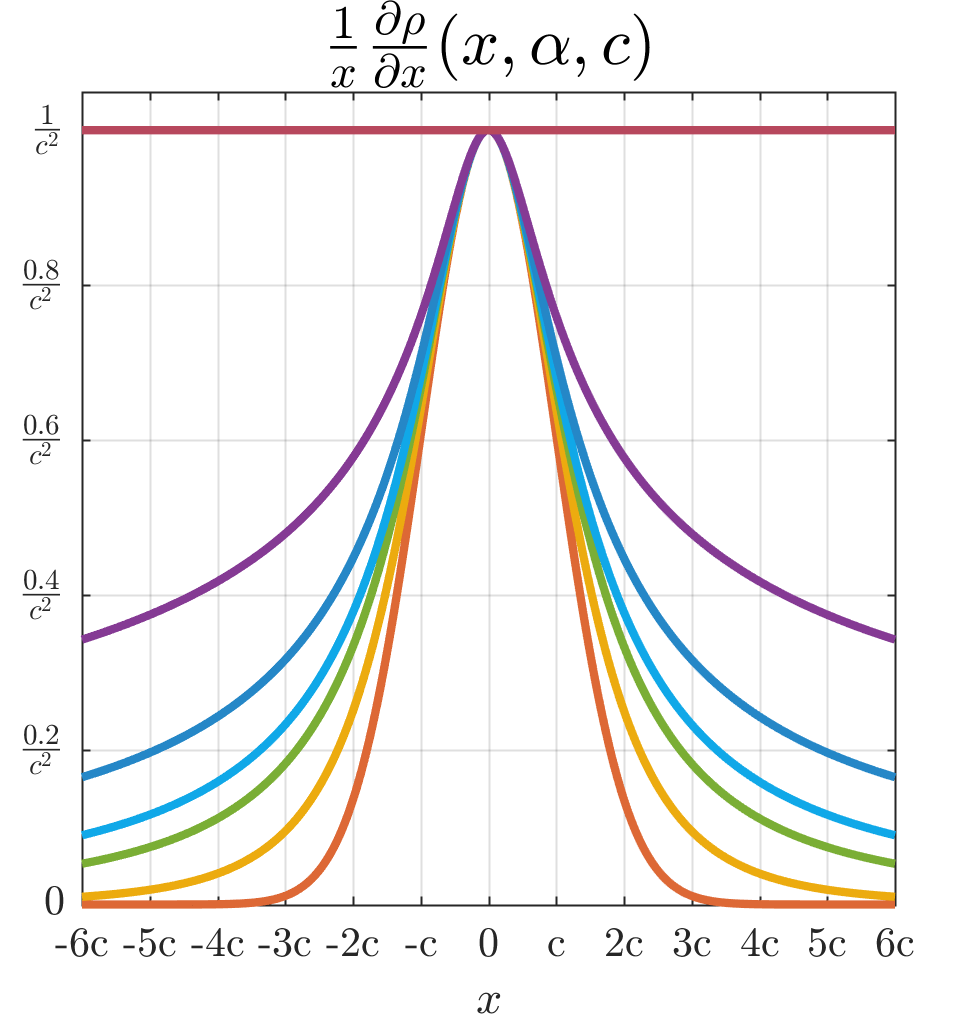} &
  \includegraphics[width=0.5\linewidth]{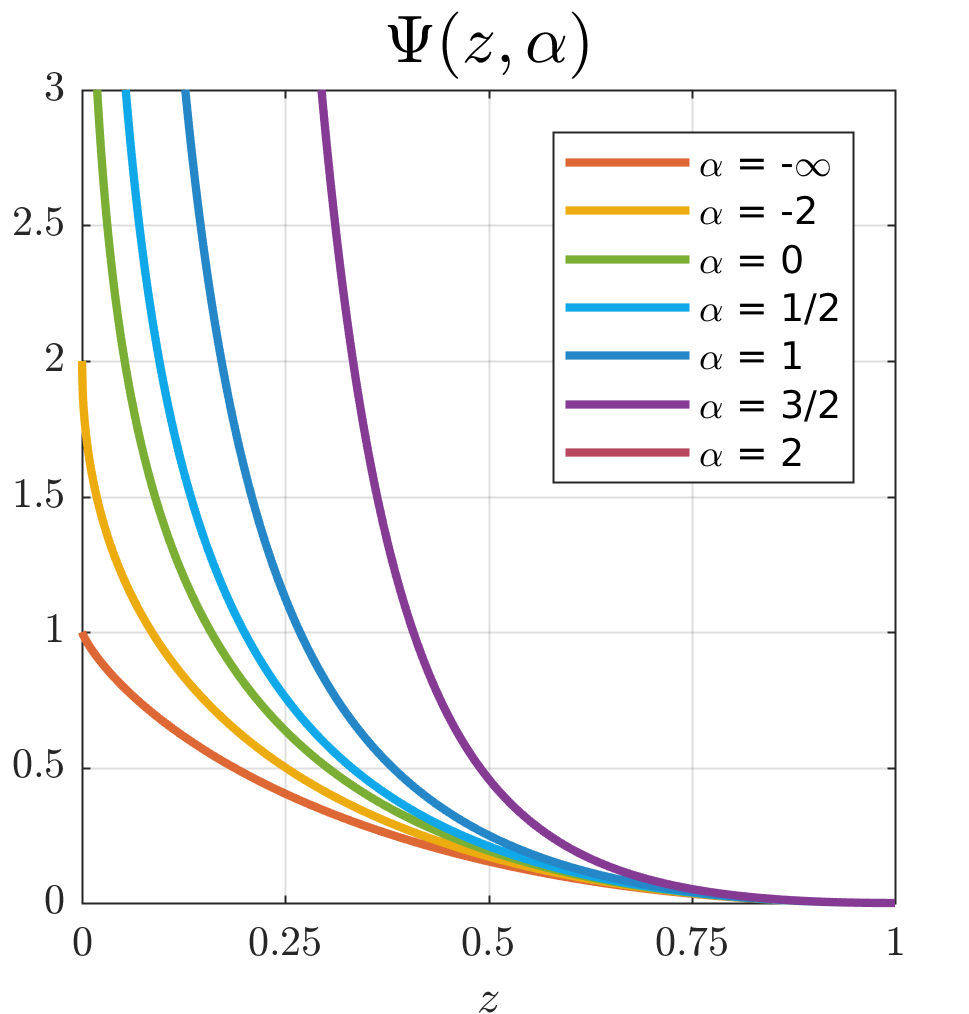}
\end{tabular}
  \caption{
   Our general loss's IRLS weight function (left) and $\Psi$-function (right)
   for different values of the shape parameter $\power$.
  \label{fig:weight_and_psi}
  }
\end{figure}

We can also derive our loss's weight function to be used during iteratively reweighted least squares \cite{IRLS, HampelEtal86}:
\begin{equation}
{1 \over x} { \deriv \rho \over \deriv x } \left(x, \power, c\right) = \begin{cases}
{1 \over c^2} & \text{if } \power = 2 \\
{2 \over x^2 + 2 c^2 } & \text{if } \power = 0 \\
{1 \over c^2} \exp \left(- {1 \over 2} \left( \sfrac{x}{c} \right)^2 \right)  & \text{if } \power = -\infty \\
{1 \over c^2} \left( { \left( \sfrac{x}{c} \right)^2 \over \abs{\power - 2} } + 1 \right)^{(\sfrac{\power}{2} - 1)} & \text{otherwise}
\end{cases}
\end{equation}
Curiously, these IRLS weights resemble a non-normalized form of Student's $t$-distribution.
These weights are not used in any of our experiments, but they are an intuitive way to
demonstrate how reducing $\power$ attenuates the effect of outliers.
A visualization of our loss's $\Psi$-functions and weight functions for different values of $\power$ can be seen in Figure~\ref{fig:weight_and_psi}.

\section{Practical Implementation}

The special cases in the definition of $\lossfun{\cdot}$
that are required because of the removable singularities of $\rawlossfun{\cdot}$ at $\power=0$ and $\power=2$
can make implementing our loss somewhat inconvenient.
Additionally, $\rawlossfun{\cdot}$ is numerically
unstable near these singularities, due to divisions by small values.
Furthermore, many deep learning frameworks handle special cases inefficiently by
evaluating all cases of a conditional statement, even though only one case is needed.
To circumvent these issues we can slightly modify our
loss (and its gradient and $\Psi$-function) to guard against singularities and
make implementation easier:
\begin{minipage}{\linewidth}
\begin{align*}
\lossfun{x, \power, c} =& {b \over d} \left( \left( {\left( \sfrac{x}{c} \right)^2 \over b } + 1 \right)^{(\sfrac{d}{2})} - 1 \right) \\
{ \deriv \rho \over \deriv x } \left(x, \power, c\right) =& {x \over c^2} \left( { \left( \sfrac{x}{c} \right)^2 \over b } + 1 \right)^{(\sfrac{d}{2} - 1)} \nonumber \\
\Psi(z, \power) = & \begin{cases}
  {b \over d} \left( \left( 1 - {d \over 2} \right) z^{d \over (d-2)} + {d z \over 2} - 1 \right)  & \text{if } \power < 2 \\
  0  & \text{if } \power = 2
\end{cases} \nonumber \\
b =& \abs{\power - 2} + \epsilon \quad\quad
d = \begin{cases} \power + \epsilon  & \text{if } \power \geq 0 \\
                   \power - \epsilon  & \text{if } \power < 0
\end{cases} \nonumber
\end{align*}
\end{minipage}
Where $\epsilon$ is some small value, such as $10^{-5}$.
Note that even very small values of $\epsilon$ can cause significant inaccuracy between
our true partition function $\partition{\power}$ and the effective partition function
of our approximate distribution when $\power$ is near $0$, so this approximate implementation
should be avoided when accurate values of $\partition{\power}$ are necessary.

\section{Partition Function Approximation}
\label{app:partition}

Implementing the negative log-likelihood of our general distribution (ie, our adaptive loss)
requires a tractable and differentiable approximation of its log partition function.
Because the analytical form of $\partition{\power}$ detailed in the paper is difficult to evaluate
efficiently for any real number, and especially difficult to differentiate with
respect to $\power$, we approximate $\log(\partition{\power})$ using cubic hermite spline interpolation in a transformed space.
Efficiently approximating $\log(\partition{\power})$ with a spline is difficult, as we would like a
concise approximation that holds over the entire valid range $\power \geq 0$, and we would like to allocate more
precision in our spline interpolation to values near $\power=2$ (which is where
$\log(\partition{\power})$ varies most rapidly).
To accomplish this, we first apply a monotonic nonlinearity to $\power$ that
stretches values near $\power=2$ (thereby increasing the density of spline knots in this region)
and compresses values as $\power \gg 4$, for which we use:
\begin{equation}
\operatorname{curve}(\power) = \begin{cases}
\frac{9 (\power - 2)}{4|\power - 2| + 1} + \power + 2  & \text{if } \power < 4 \\
              \frac{5}{18} \log \left(4 \power - 15 \right) + 8  & \text{otherwise}
\end{cases}
\end{equation}
This curve is roughly piecewise-linear in $[0, 4]$ with a slope of $\sim\!\!\!1$ at $\power=0$
and $\power=4$, but with a slope of $\sim\!\!\!10$ at $\power=2$. When $\power > 4$ the
curve becomes logarithmic.
This function is continuously differentiable, as is required for our log-partition
approximation to also be continuously differentiable.

\begin{figure}[t]
\centering
  \includegraphics[width=\linewidth]{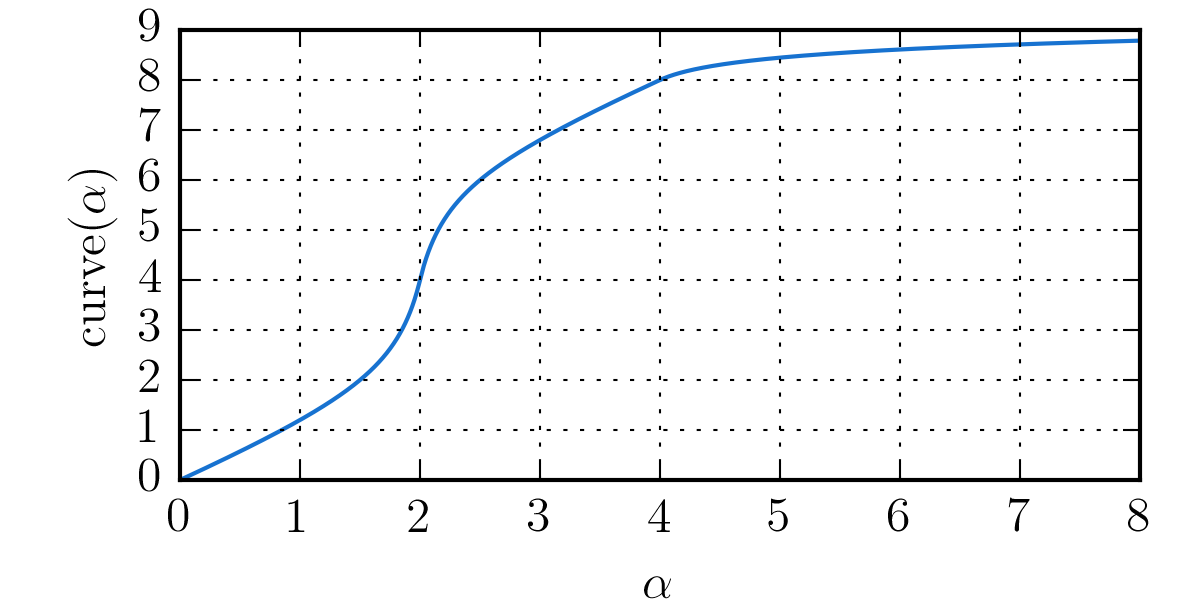}
  \includegraphics[width=\linewidth]{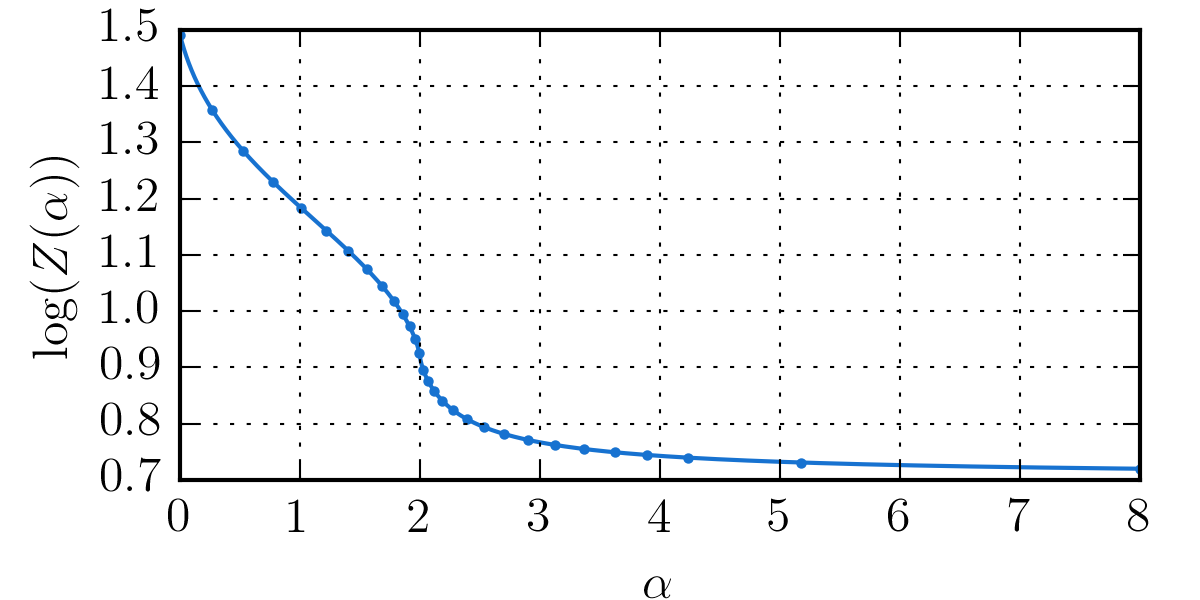}
  \caption{
  Because our distribution's log partition function $\log(\partition{\power})$ is
  difficult to evaluate for arbitrary inputs, we approximate it using
  cubic hermite spline interpolation in a transformed space: first we curve $\power$ by a continuously differentiable
  nonlinearity that increases knot density near $\power=2$ and decreases knot density when $\power>4$ (top)
  and then we fit an evenly-sampled cubic hermite spline in that curved space (bottom).
  The dots shown in the bottom plot are a subset of the knots used by our cubic spline, and
  are presented here to demonstrate how this approach allocates spline knots with respect to $\power$.
  \label{fig:partition}
  }
\end{figure}

We transform $\power$ with this nonlinearity, and then approximate
$\log(\partition{\power})$ in that transformed space using a spline
with knots in the range of $[0, 12]$ evenly spaced apart by $\sfrac{1}{1024}$.
Values for each knot are set to their true value, and tangents for
each knot are set to minimize the squared error between the spline
and the true log partition function.
Because our spline knots are evenly spaced in this transformed space, spline
interpolation can be performed in constant time with respect to the number of spline knots.
For all values of $\power$ this approximation is accurate to within  $10^{-6}$, which appears to be sufficient for our purposes.
Our nonlinearity and our spline approximation to the true partition function for small values of $\power$ can be seen in Figure~\ref{fig:partition}.

\section{Motivation and Derivation}

Our loss function is derived from the ``generalized Charbonnier'' loss \cite{SunCVPR10},
which itself builds upon the Charbonnier loss function \cite{Charb1994}.
To better motivate the construction of our loss function, and to clarify its
relationship to prior work, here we work through how our loss function was
constructed.

Generalized Charbonnier loss can be defined as:
\begin{equation}
\charb{x, \power, c} = \left( x^2 + c^2 \right)^{\sfrac{\power}{2}}
\end{equation}
Here we use a slightly different parametrization from \cite{SunCVPR10} and use ${\sfrac{\power}{2}}$ as the exponent instead of just $\power$.
This makes the generalized Charbonnier somewhat easier to reason about with respect to standard loss functions:
$\charb{x, 2, c}$ resembles L2 loss,
$\charb{x, 1, c}$ resembles L1 loss, etc.

We can reparametrize generalized Charbonnier loss as:
\begin{equation}
\charb{x, \power, c} = c^{\power} \left( \left( \sfrac{x}{c} \right)^2 + 1 \right)^{\sfrac{\power}{2}}
\end{equation}
We omit the $c^{\power}$ scale factor, which gives us a loss that is scale invariant with respect to $c$:
\begin{align}
\gencharb{x, \power, c} = \left( (\sfrac{x}{c})^2 + 1 \right)^{\sfrac{\power}{2}} \label{eq:traditional} \\
\forall_{k>0} \quad g(k x, \power, k c) = g(x, \power, c) \label{eq:scalinv}
\end{align}
This lets us view the $c$ ``padding'' variable as a ``scale'' parameter,
similar to other common robust loss functions.
Additionally, only after dropping this scale factor does setting $\power$ to a
negative value yield a family of meaningful robust loss functions, such as
Geman-McClure loss.

But this loss function still has several unintuitive properties:
the loss is non-zero when $x=0$ (assuming a non-zero value of $c$),
and the curvature of the quadratic ``bowl'' near $x=0$ varies as a function of $c$ and $\power$.
We therefore construct a shifted and scaled version of Equation~\ref{eq:traditional} that does
not have these properties:
\begin{equation}
{\gencharb{x, \power, c} - \gencharb{0, \power, c} \over c^2 g'' \left( 0, \power, c \right) }
= {1 \over \power} \left( \left( \left(\sfrac{x}{c} \right)^2 + 1 \right)^{\sfrac{a}{2}} - 1 \right)
\end{equation}
This loss generalizes L2, Cauchy, and Geman-McClure loss, but it has the
unfortunate side-effect of flattening out to $0$ when $\power \ll 0$,
thereby prohibiting many annealing strategies.
This can be addressed by modifying the $1/\power$ scaling to approach $1$ instead of $0$
when $\power \ll 0$
by introducing another scaling that cancels out the division by $\power$.
To preserve the scale-invariance of Equation~\ref{eq:scalinv},
this scaling also needs to be applied to the $\left( \sfrac{x}{c} \right)^2$
term in the loss.
This scaling also needs to maintain the monotonicity of our loss with respect to $\power$
so as to make annealing possible.
There are several scalings that satisfy this property, so we
select one that is efficient to evaluation and which keeps our loss function smooth
(ie, having derivatives of all orders everywhere) with respect to $x$, $\power$, and $c$, which is
$\abs{\power - 2}$. This gives us our final loss function:
\begin{equation}
\rawlossfun{x, \power, c} = {\abs{\power - 2} \over \power } \left( \left( {\left( \sfrac{x}{c} \right)^2 \over \abs{\power - 2} } + 1 \right)^{\sfrac{\power}{2}} - 1 \right)
\end{equation}
Using $\abs{\power - 2}$ satisfies all of our criteria, though
it does introduce a removable singularity into our loss function at $\power=2$ and
reduces numerical stability near $\power=2$.

\section{Additional Properties}
\label{app:additional_properties}
Here we enumerate additional properties of our loss function that were not
used in our experiments.

At the origin the IRLS weight of our loss is $\frac{1}{c^2}$:
\begin{equation}
{1 \over x} { \deriv \rho \over \deriv x } \left(0, \power, c\right) = {1 \over c^2}
\end{equation}
For all values of $\power$, when $\abs{x}$ is small with respect to $c$ the loss is well-approximated by a quadratic bowl:
\begin{equation}
\lossfun{x, \power, c} \approx {1 \over 2} \left( \sfrac{x}{c} \right)^2 \quad \text{if } |x| < c
\end{equation}
Because the second derivative of the loss is maximized at $x = 0$, this quadratic
approximation tells us that the second derivative is bounded from above:
\begin{equation}
{\deriv^2 \rho \over \deriv x^2} (x, \power, c) \leq {1 \over c^2}
\end{equation}
When $\power$ is negative the loss approaches a constant as $|x|$ approaches infinity,
letting us bound the loss:
\begin{equation}
\forall_{x, c} \,\, \lossfun{x, \power, c} \leq { \power - 2 \over \power } \quad \text{if } \power < 0
\end{equation}
The loss's $\Psi$-function increases monotonically with respect to $\power$ when $\power < 2$ for all values of $z$ in $[0, 1]$:
\begin{equation}
{ \deriv \Psi \over \deriv \power } \left(z, \power \right) \geq 0 \quad \text{if } 0 \leq z \leq 1
\end{equation}

The roots of the second derivative of $\lossfun{x, \power, c}$ are:
\begin{equation}
x = \pm c \sqrt{\power - 2 \over \power - 1}
\end{equation}
This tells us at what value of $x$ the loss begins to redescend.
This point has a magnitude of $c$ when $\power = -\infty$, and that magnitude increases as $\power$ increases.
The root is undefined when $\power \geq 1$, as our loss is redescending iff $\power < 1$.
Our loss is strictly convex iff $\power \geq 1$, non-convex iff $\power < 1$, and pseudoconvex for all values of $\power$.

\section{Wavelet Implementation}
\label{sec:wavelet}

Two of our experiments impose our loss on images reparametrized with the
Cohen-Daubechies-Feauveau (CDF) 9/7 wavelet decomposition \cite{cohen1992biorthogonal}.
The analysis filters used for these experiments are:
\begin{equation}
\begin{tabular}{c | c}
 lowpass & highpass \\ \hline
0.852698679009 & 0.788485616406\\
0.377402855613 & -0.418092273222\\
-0.110624404418 & -0.040689417609\\
-0.023849465020 & 0.064538882629 \\
0.037828455507 & \\
\end{tabular} \nonumber
\end{equation}
Here the origin coefficient of the filter is listed first, and the
rest of the filter is symmetric.
The synthesis filters are defined as usual, by reversing the sign of alternating
wavelet coefficients in the analysis filters.
The lowpass filter sums to $\sqrt{2}$, which means that image intensities are
doubled at each scale of the wavelet decomposition, and that the magnitude of an
image is preserved in its wavelet decomposition.
Boundary conditions are ``reflecting'', or half-sample symmetric.

\section{Variational Autoencoders}
\label{app:vae}

Our VAE experiments were performed using the code
included in the TensorFlow Probability codebase at
\url{http://github.com/tensorflow/probability/blob/master/tensorflow_probability/examples/vae.py}.
This code was designed for binarized MNIST data, so adapting it to
the real-valued color images in CelebA~\cite{Liu2015} required the following changes:
\begin{itemize}[leftmargin=0.15in]
\item Changing the input and output image resolution from $(28, 28, 1)$ to $(64, 64, 3)$.
\item Increasing the number of training steps from 5000 to 50000, as CelebA is significantly larger than MNIST.
\item Delaying the start of cosine decay of the learning rate until the final 10000 training iterations.
\item Changing the CNN architecture from a 5-layer network with 5-tap and 7-tap
      filters with interleaved strides of 1 and 2 (which maps from a $28 \times 28$ image to a vector)
      to a 6-layer network consisting of all 5-tap filters with strides of 2
      (which maps from a $64 \times 64$ input to a vector).
      The number of hidden units was left unchanged, and the one extra layer we added at the
      end of our decoder (and beginning of our decoder) was given the same number
      of hidden units as the layer before it.
\item In our ``DCT + YUV'' and ``Wavelets + YUV'' models, before imposing our model's
      posterior we apply an RGB-to-YUV transformation and then a per-channel DCT
      or wavelet transformation to the YUV images, and then invert these
      transformations to visualize each sampled image. In the ``Pixels + RGB'' model
      this transformation and its inverse are the identity function.
\item As discussed in the paper, for each output coefficient (pixel value, DCT coefficient, or wavelet coefficient)
we add a scale variable ($\sigma$ when using normal distributions, $c$ when using our general distributions)
and a shape variable $\power$ (when using our general distribution).
\end{itemize}
We made as few changes to the reference code as possible so as to
keep our model architecture as simple as possible, as our goal is not to produce
state-of-the-art image synthesis results for some task, but is instead to simply demonstrate
the value of our general distribution in isolation.

CelebA~\cite{Liu2015} images are processed by extracting a square $160 \times 160$
image region at the center of each $178 \times 218$ image and downsampling it
to $64 \times 64$ by a factor of $2.5 \times$ using TensorFlow's
bilinear interpolation implementation. Pixel intensities are scaled to $[0, 1]$.

\begin{figure}  \centering
  \includegraphics[width=2.7in]{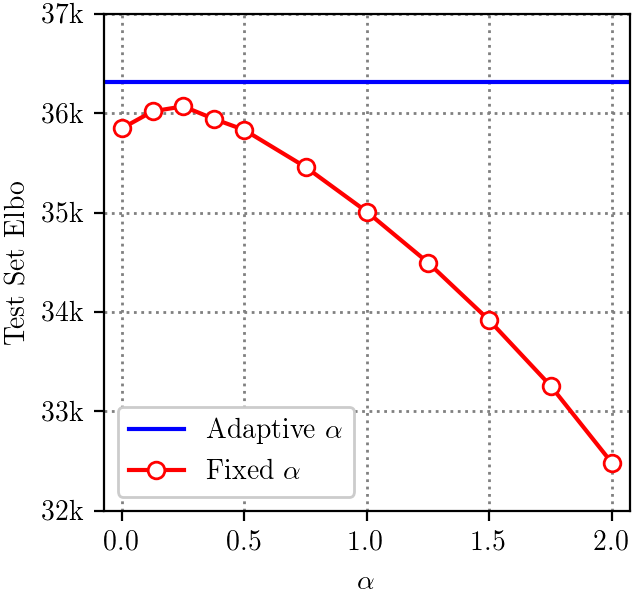}
  \caption{Here we compare the validation set ELBO of our adaptive ``Wavelets + YUV''
  VAE model with the ELBO achieved when setting all wavelet coefficients
  to have the same fixed shape parameter $\power$. We see that allowing our
  distribution to individually adapt its shape parameter to each coefficient
  outperforms any single fixed shape parameter.
  \label{fig:fixed_sweep}
  }
\end{figure}

In the main paper we demonstrated that using our general distribution to independently
model the robustness of each coefficient of our image representation works better
than assuming a Cauchy ($\power=0$) or normal distribution ($\power=2$) for all coefficients
(as those two distributions lie within our general distribution). To
further demonstrate the value of \emph{independently} modeling the robustness
of each individual coefficient, we ran a more thorough experiment in which we
densely sampled values for $\power$ in $[0, 2]$ that are used for \emph{all} coefficients.
In Figure~\ref{fig:fixed_sweep} we visualize the validation set ELBO for each fixed
value of $\power$ compared to an independently-adapted model.
As we can see, though quality can be improved by selecting a value for $\power$ in
between $0$ and $2$, no single global setting of the shape parameter matches
the performance achieved by allowing each coefficient's shape parameter to
automatically adapt itself to the training data. This observation is consistent with
earlier results on adaptive heavy-tailed distributions for image data~\cite{PortillaSWS03}.

In our Student's t-distribution experiments, we parametrize each ``degrees of freedom''
parameter as the exponentiation of some latent free parameter:
\begin{equation}
\nu^{(i)} = \exp \left(\nu_\ell^{(i)} \right)
\end{equation}
where all $\nu_\ell^{(i)}$ are initialized to $0$. Technically, these experiments
are performed with the ``Generalized Student's t-distribution'', meaning that we have an additional
scale parameter $\sigma^{(i)}$ that is divided into $x$ before computing the
log-likelihood and is accounted for in the partition function.
These scale parameters are parametrized identically to the $c^{(i)}$ parameters
used by our general distribution.

Comparing likelihoods across our different image representations requires
that the ``Wavelets + YUV'' and ``DCT + YUV'' representations be normalized to match
the ``Pixels + RGB'' representation. We therefore construct the linear transformations used
for the ``Wavelets + YUV'' and ``DCT + YUV'' spaces to have determinants of $1$ as per the change of variable formula
(that is, both transformations are in the ``special linear group'').
Our wavelet construction in Section~\ref{sec:wavelet} satisfies this criteria, and we use the
orthonormal version of the DCT which also satisfies this criteria. However, the standard RGB to
YUV conversion matrix does not have a determinant of $1$, so we scale it by the inverse
of the cube root of the standard conversion matrix, thereby forcing its determinant to be $1$.
The resulting matrix is:
\[
\begin{bmatrix}
    \phantom{-}0.47249 & \phantom{-}0.92759 & \phantom{-}0.18015 \\
    -0.23252 & -0.45648 & \phantom{-}0.68900 \\
    \phantom{-}0.97180 & -0.81376 & -0.15804
\end{bmatrix}
\]
Naturally, its inverse maps from YUV to RGB.

\begin{figure}[t]
  \centering
  \begin{tabular}{@{}c@{\hspace{0.05in}}c@{\hspace{0.05in}}c@{\hspace{0.05in}}c@{\hspace{0.05in}}@{}}
  \rotatebox[origin=l]{90}{\quad\quad\quad$\{ \power^{(i)} \}$}&
  \includegraphics[width=0.95in]{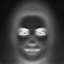} &
  \includegraphics[width=0.95in]{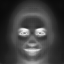} &
  \includegraphics[width=0.95in]{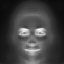} \\
  \rotatebox[origin=l]{90}{\quad\quad$ \{ \log \left(c^{(i)} \right) \} $}&
  \includegraphics[width=0.95in]{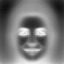} &
  \includegraphics[width=0.95in]{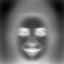} &
  \includegraphics[width=0.95in]{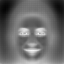} \\
  & R & G & B
  \end{tabular}
  \caption{
  The final shape and scale parameters $\{ \power^{(i)} \}$ and $\{ c^{(i)} \}$
  for our ``Pixels + RGB'' VAE after training has converged.
  We visualize $\power$ with black=$0$ and white=$2$ and $\log(c)$ with black=$\log(0.002)$ and white=$\log(0.02)$.
  \label{fig:vae_rgb_loss}
  }
  \vspace{0.2in}
  \centering
  \begin{tabular}{@{}c@{\hspace{0.05in}}c@{\hspace{0.05in}}c@{\hspace{0.05in}}c@{\hspace{0.05in}}@{}}
  \rotatebox[origin=l]{90}{\quad\quad\quad$\{ \power^{(i)} \}$}&
  \includegraphics[width=0.95in]{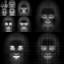} &
  \includegraphics[width=0.95in]{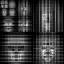} &
  \includegraphics[width=0.95in]{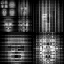} \\
  \rotatebox[origin=l]{90}{\quad\quad$ \{ \log \left(c^{(i)} \right) \} $}&
  \includegraphics[width=0.95in]{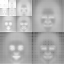} &
  \includegraphics[width=0.95in]{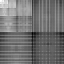} &
  \includegraphics[width=0.95in]{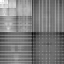} \\
  & Y & U & V
  \end{tabular}
  \caption{
  The final shape and scale parameters $\{ \power^{(i)} \}$ and $\{ c^{(i)} \}$
  for our ``Wavelets + YUV'' VAE after training has converged.
  We visualize $\power$ with black=$0$ and white=$2$ and $\log(c)$ with black=$\log(0.00002)$ and white=$\log(0.2)$.
  \label{fig:vae_wavelet_loss}
  }
\end{figure}

Because our model can adapt the shape and scale
parameters of our general distribution to each output coefficient,
after training we can inspect the shapes and scales that
have emerged during training, and from them gain insight into how optimization has
modeled our training data. In Figures~\ref{fig:vae_rgb_loss}
and \ref{fig:vae_wavelet_loss} we visualize the shape and scale parameters for
our ``Pixels + RGB'' and ``Wavelets + YUV'' VAEs respectively. Our ``Pixels''
model is easy to visualize as each output coefficient simply corresponds to
a pixel in a channel, and our ``Wavelets'' model can be visualized by flattening
each wavelet scale and orientation into an image (our DCT-based model is difficult to
visualize in any intuitive way). In both models we observe that training
has determined that these face images should be modeled using normal-like
distributions near the eyes and mouth, presumably because these structures are
consistent and repeatable on human faces, and Cauchy-like distributions on the
background and in flat regions of skin.
Though our ``Pixels + RGB'' model has estimated similar
distributions for each color channel, our ``Wavelets + YUV'' model has estimated
very different behavior for luma and chroma: more Cauchy-like behavior is expected in
luma variation, especially at fine frequencies, while chroma variation is modeled
as being closer to a normal distribution across all scales.

\newcommand{\noisewidth}{0.45\linewidth}
\begin{figure}  \centering
  \begin{tabular}{@{}c@{\hspace{0.08in}}c@{\hspace{0.08in}}c@{}}
  & Mean Reconstruction & Sampled Reconstruction \\
  \rotatebox[origin=l]{90}{\quad\quad\, Pixels + RGB} &
  \includegraphics[width=\noisewidth]{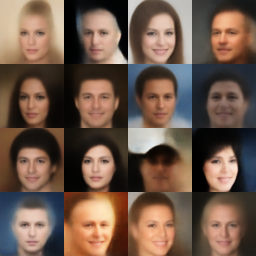} &
  \includegraphics[width=\noisewidth]{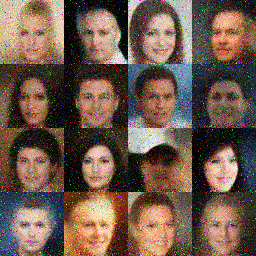} \\
  \rotatebox[origin=l]{90}{\quad\quad\, DCT + YUV} &
  \includegraphics[width=\noisewidth]{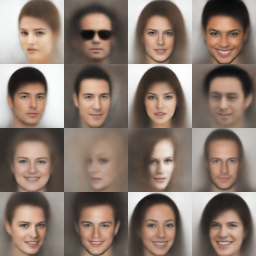} &
  \includegraphics[width=\noisewidth]{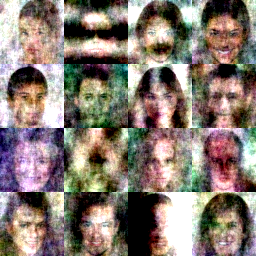} \\
  \rotatebox[origin=l]{90}{\quad\,\,\, Wavelets + YUV} &
  \includegraphics[width=\noisewidth]{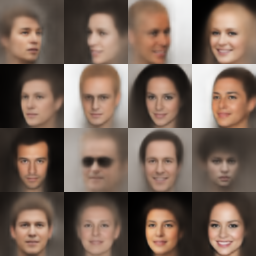} &
  \includegraphics[width=\noisewidth]{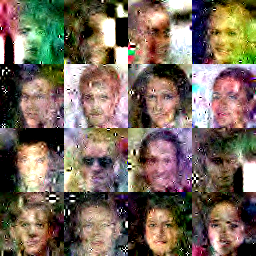}
  \end{tabular}
  \caption{
  As is common practice, the VAE samples shown in this paper are samples from
  the latent space (left) but not from the final conditional distribution (right).
  Here we contrast decoded means and samples from VAEs using our different output
  spaces, all using our general distribution.
  \label{fig:noise}
  }
\end{figure}

See Figure~\ref{fig:big_vae_samples} for additional samples from our models,
and see Figure~\ref{fig:big_vae_recon} for reconstructions from our models on
validation-set images.
As is common practice, the samples and reconstructions in those figures and in the paper
are the means of the output distributions of the decoder, not samples from those distributions.
That is, we draw samples from the latent encoded space
and then decode them, but we do not draw samples in our output space.
Samples drawn from these output distributions tend to look noisy
and irregular across all distributions and image representations, but they provide
a good intuition of how our general distribution behaves in each image representation,
so in Figure~\ref{fig:noise} we present side-by-side visualizations of decoded means and samples.

\section{Unsupervised Monocular Depth Estimation}
\label{app:monodepth}

Our unsupervised monocular depth estimation experiments use the code from
\url{https://github.com/tinghuiz/SfMLearner},
which appears to correspond to the ``Ours (w/o explainability)'' model
from Table 1 of \cite{zhou2017}.
The only changes we made to this code were: replacing its loss function with our own,
reducing the number of training iterations from $200000$ to $100000$ (training
converges faster when using our loss function) and disabling the
smoothness term and multi-scale side predictions used by \cite{zhou2017}, as neither
yielded much benefit when combined with our new loss function
and they complicated experimentation by introducing hyperparameters.
Because the reconstruction loss in \cite{zhou2017} is
the sum of the means of the losses imposed at each scale in a $D$-level pyramid
of side predictions,
we use a $D$ level normalized wavelet decomposition
(wherein images in $[0,1]$ result in wavelet coefficients in $[0,1]$) and then scale each coefficient's
loss by $2^d$, where $d$ is the coefficient’s level.

\begin{figure}
  \centering
  \begin{tabular}{@{}c@{\hspace{0.05in}}c@{}}
  \rotatebox[origin=l]{90}{\quad\quad\,\, Y} & \includegraphics[width=3in]{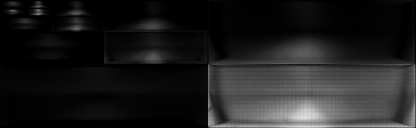} \\
  \rotatebox[origin=l]{90}{\quad\quad\,\, U} & \includegraphics[width=3in]{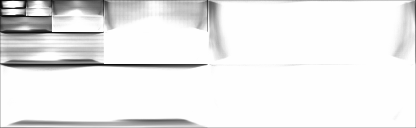} \\
  \rotatebox[origin=l]{90}{\quad\quad\,\, V} & \includegraphics[width=3in]{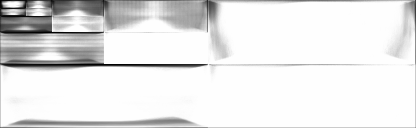}
  \end{tabular}
  \caption{
  The final shape parameters $\power$ for our unsupervised monocular depth
  estimation model trained on KITTI data.
  The parameters are visualized in the same ``YUV + Wavelet'' output space as was
  used during training, where black is $\power=0$ and white is $\power=2$.
  \label{fig:sfm_loss}
  }
\end{figure}

In Figure~\ref{fig:sfm_loss} we visualize the final shape parameters
for each output coefficient that were converged upon during training.
These results provide some
insight into why our adaptive model produces improved results compared to the
ablations of our model in which we use a single fixed or annealed value for $\power$
for all output coefficients.
From the low $\power$ values in the luma channel
we can infer that training has decided that luma variation often has outliers,
and from the high $\power$ values in the chroma channel
we can infer that chroma variation rarely has outliers.
Horizontal luma variation (upper right) tends to have larger $\power$ values
than vertical luma variation (lower left), perhaps because
depth in this dataset is largely due to horizontal motion, and so horizontal
gradients tend to provide more depth information than vertical gradients.
Looking at the sides and the bottom of all scales and channels we see that the model expects more
outliers in these regions, which is likely due to boundary effects: these areas often contain
consistent errors due to there not being a matching pixel in the alternate view.

In Figures~\ref{fig:sfm_results_big1} and \ref{fig:sfm_results_big2} we present
many more results from the test split of the KITTI dataset, in which we compare
our ``adaptive'' model's output to the baseline model and the ground-truth depth.
The improvement we see is substantial and consistent across a variety of scenes.

\section{Fast Global Registration}

Our registration results were produced using the code release corresponding to \cite{Zhou16}.
Because the numbers presented in \cite{Zhou16} have low precision,
we reproduced the performance of the baseline FGR algorithm using this code.
This code included some evaluation details that were omitted from the paper that
we determined through correspondence with the author:
for each input, FGR is run $20$ times with random initialization and the median error is reported.
We use this procedure when reproducing the baseline performance of \cite{Zhou16}
and when evaluating our own models.

\renewcommand{\vaewidth}{0.3\linewidth}
\begin{figure*}[p]
\centering
  \begin{tabular}{l@{\hspace{0.05in}}c@{\hspace{0.05in}}c@{\hspace{0.05in}}c}
    &
    Pixels + RGB
    &
    DCT + YUV
    &
    Wavelets + YUV
    \\
    \rotatebox[origin=l]{90}{\quad\quad\quad Normal distribution}
    &
    \includegraphics[width=\vaewidth]{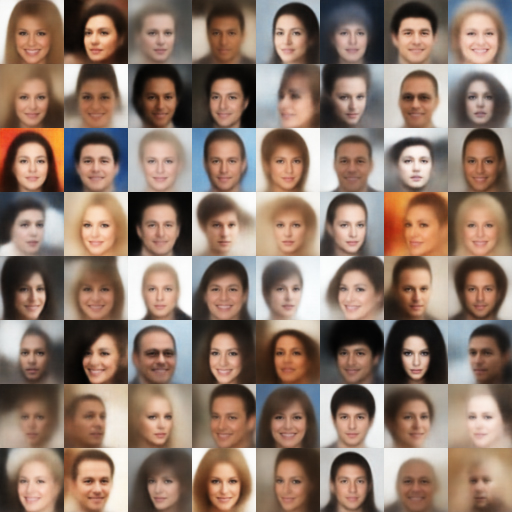}
    &
    \includegraphics[width=\vaewidth]{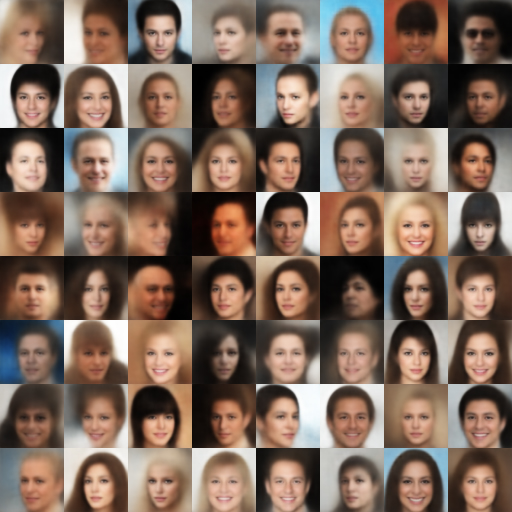}
    &
    \includegraphics[width=\vaewidth]{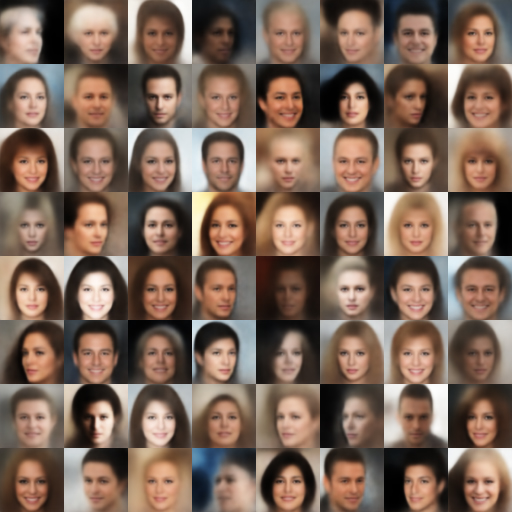}
    \\
    \rotatebox[origin=l]{90}{\quad\quad\quad Cauchy distribution}
    &
    \includegraphics[width=\vaewidth]{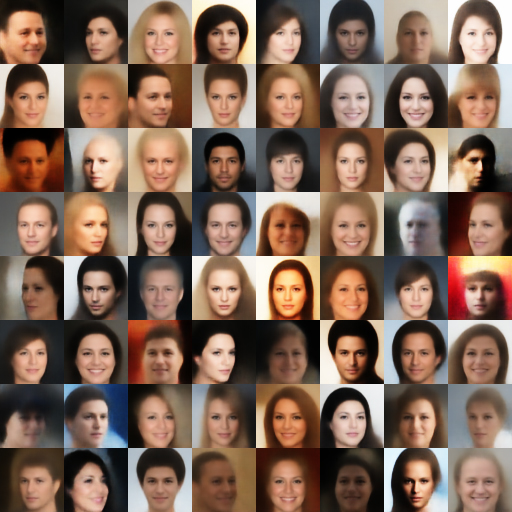}
    &
    \includegraphics[width=\vaewidth]{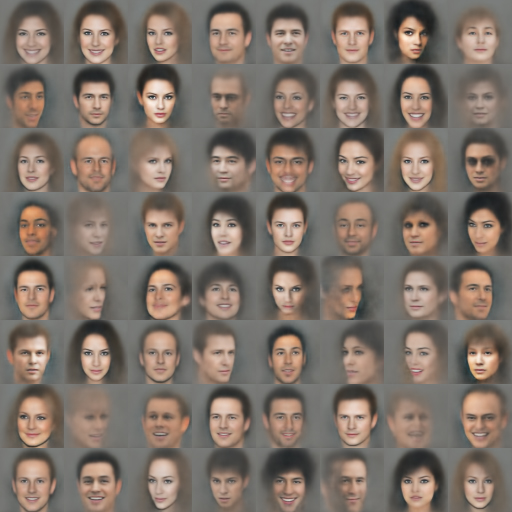}
    &
    \includegraphics[width=\vaewidth]{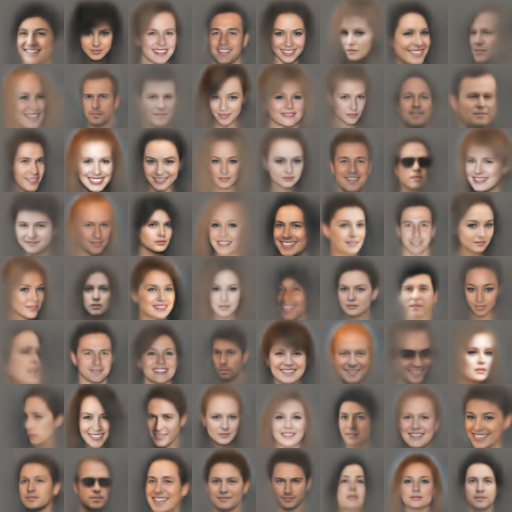}
    \\
    \rotatebox[origin=l]{90}{\quad\quad\quad Student's t-distribution}
    &
    \includegraphics[width=\vaewidth]{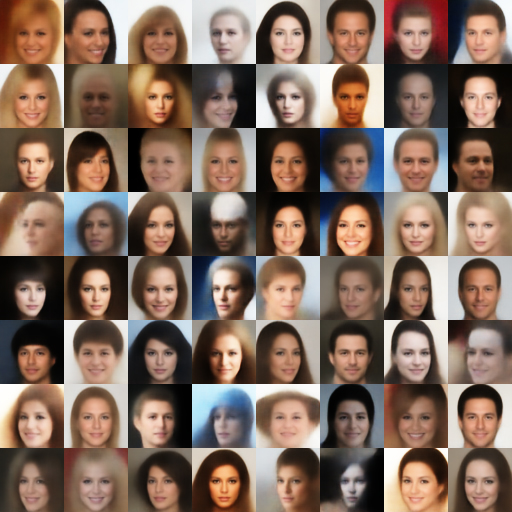}
    &
    \includegraphics[width=\vaewidth]{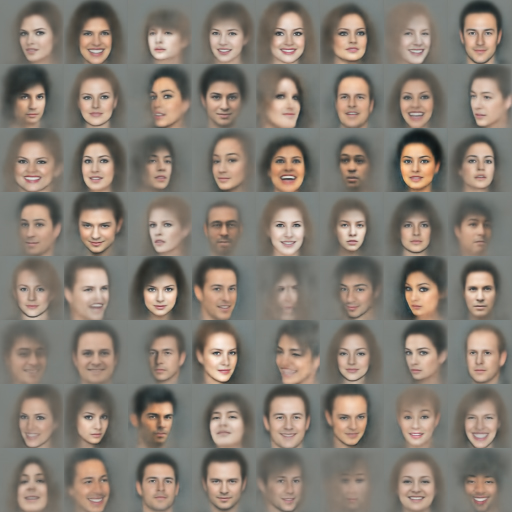}
    &
    \includegraphics[width=\vaewidth]{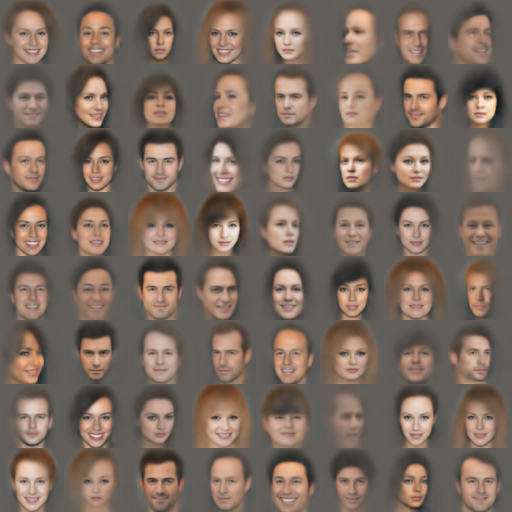}
    \\
    \rotatebox[origin=l]{90}{\quad\quad\quad\quad Our distribution}
    &
    \includegraphics[width=\vaewidth]{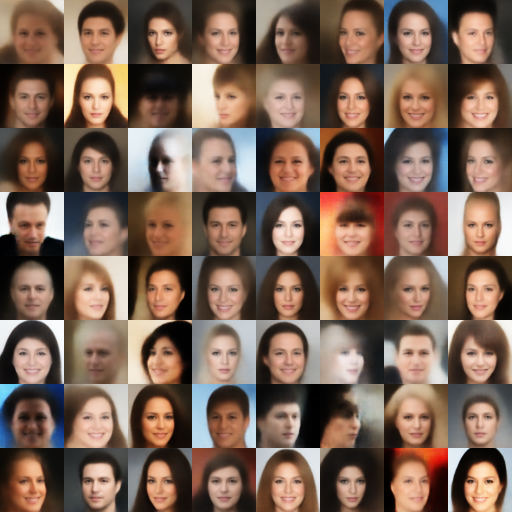}
    &
    \includegraphics[width=\vaewidth]{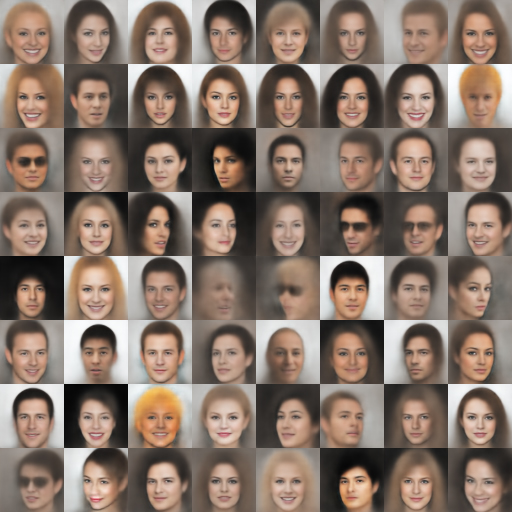}
    &
    \includegraphics[width=\vaewidth]{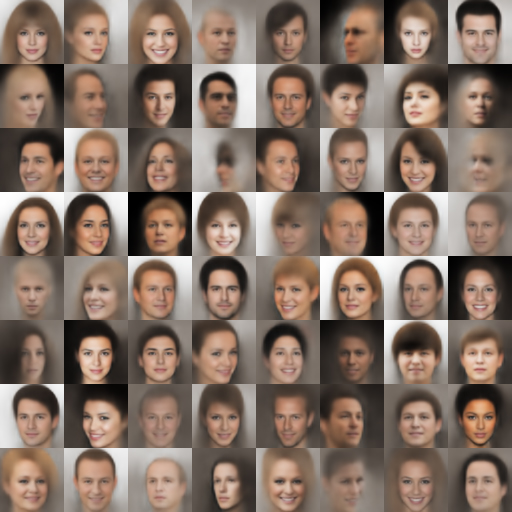}
  \end{tabular}
  \caption{
    Random samples
    (more precisely, means of the output distributions decoded from random samples in our latent space)
    from our family of trained variational autoencoders.
  \label{fig:big_vae_samples}
  }
\end{figure*}

\renewcommand{\vaewidth}{0.072\linewidth}
\begin{figure*}[p]
\centering
  \begin{tabular}{@{}c@{\hspace{0.08in}}|c@{}c@{}c@{}c@{\hspace{0.08in}}c@{}c@{}c@{}c@{\hspace{0.08in}}c@{}c@{}c@{}c@{}}
  &
  \multicolumn{4}{c}{Pixels + RGB}
  &
  \multicolumn{4}{c}{DCT + YUV}
  &
  \multicolumn{4}{c}{Wavelets + YUV} \\
  Input &
  Normal &
  Cauchy &
  t-dist &
  Ours &
  Normal &
  Cauchy &
  t-dist &
  Ours &
  Normal &
  Cauchy &
  t-dist &
  Ours \\
  \includegraphics[width=\vaewidth]{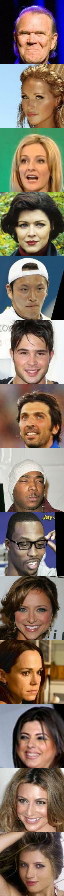} &
  \includegraphics[width=\vaewidth]{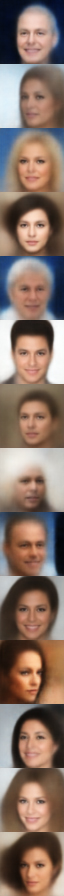} &
  \includegraphics[width=\vaewidth]{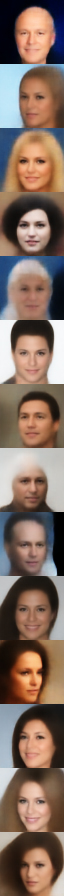} &
  \includegraphics[width=\vaewidth]{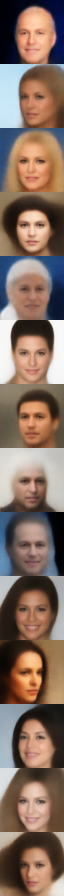} &
  \includegraphics[width=\vaewidth]{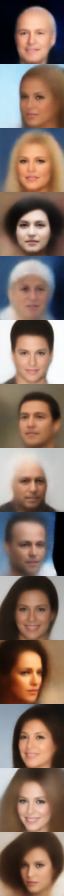} &
  \includegraphics[width=\vaewidth]{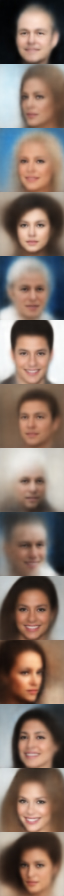} &
  \includegraphics[width=\vaewidth]{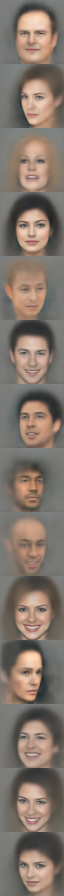} &
  \includegraphics[width=\vaewidth]{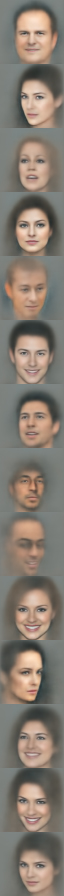} &
  \includegraphics[width=\vaewidth]{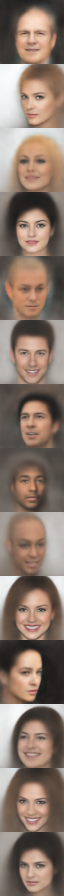} &
  \includegraphics[width=\vaewidth]{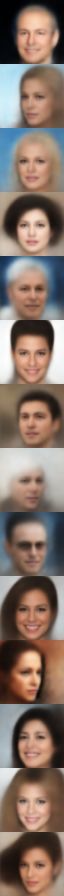} &
  \includegraphics[width=\vaewidth]{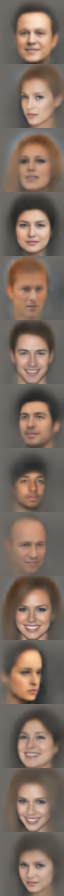} &
  \includegraphics[width=\vaewidth]{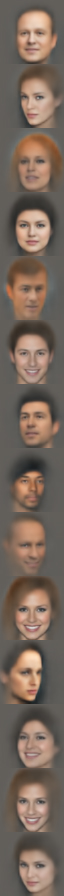} &
  \includegraphics[width=\vaewidth]{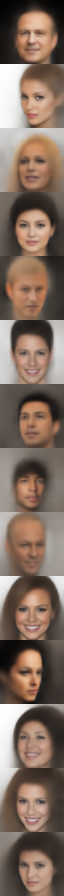}
  \end{tabular}
  \caption{
  Reconstructions from our family of trained variational autoencoders, in which
  we use one of three different image representations for modeling
  images (super-columns) and use either
    normal, Cauchy, Student's t, or our general distributions for modeling the coefficients of each
    representation (sub-columns).
  The leftmost column shows the images which are used as input to each autoencoder.
  Reconstructions from models using general
  distributions tend to be sharper and more detailed than reconstructions from
  the corresponding model that uses normal distributions, particularly for the
  DCT or wavelet representations, though this difference is less pronounced
  than what is seen when comparing samples from these models. The DCT and
  wavelet models trained with Cauchy distributions or Student's t-distributions systematically fail to
  preserve the background of the input image, as was noted when observing samples
  from these distributions.
  \label{fig:big_vae_recon}
  }
\end{figure*}

\renewcommand{\sfmwidth}{0.41\linewidth}
\renewcommand{\nametag}{000}

\begin{figure*}[p]
\centering
  \renewcommand{\nametag}{010}
  \begin{tabular}{@{}l@{\hspace{0.04in}}c@{}}
    \rotatebox[origin=l]{90}{\,\,\,\,\,\,Input}
    &
    \includegraphics[width=\sfmwidth]{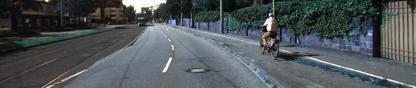}
    \\
    \rotatebox[origin=l]{90}{\,Baseline}
    &
    \includegraphics[width=\sfmwidth]{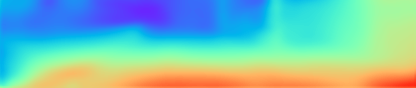}
    \\
    \rotatebox[origin=l]{90}{\,\,\,\,\,\,Ours}
    &
    \includegraphics[width=\sfmwidth]{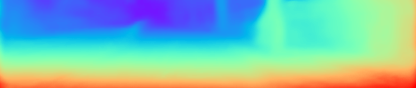}
    \\
    \rotatebox[origin=l]{90}{\,\,\,\,Truth}
    &
    \includegraphics[width=\sfmwidth]{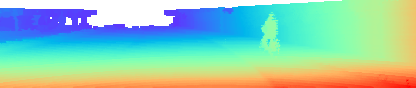}
    \\
  \end{tabular}
  \hspace{0.15in}
  \renewcommand{\nametag}{142}
  \begin{tabular}{@{}l@{\hspace{0.04in}}c@{}}
    \rotatebox[origin=l]{90}{\,\,\,\,\,\,Input}
    &
    \includegraphics[width=\sfmwidth]{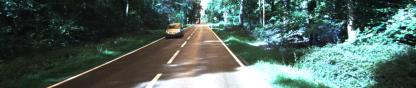}
    \\
    \rotatebox[origin=l]{90}{\,Baseline}
    &
    \includegraphics[width=\sfmwidth]{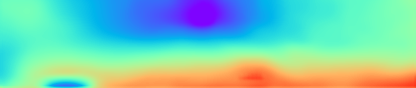}
    \\
    \rotatebox[origin=l]{90}{\,\,\,\,\,\,Ours}
    &
    \includegraphics[width=\sfmwidth]{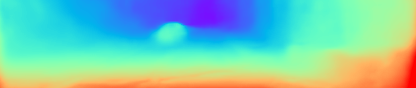}
    \\
    \rotatebox[origin=l]{90}{\,\,\,\,Truth}
    &
    \includegraphics[width=\sfmwidth]{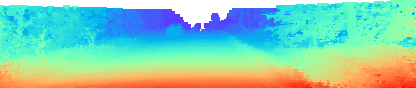}
    \\
  \end{tabular}
  \\
  \vspace{0.2in}
  \renewcommand{\nametag}{164}
  \begin{tabular}{@{}l@{\hspace{0.04in}}c@{}}
    \rotatebox[origin=l]{90}{\,\,\,\,\,\,Input}
    &
    \includegraphics[width=\sfmwidth]{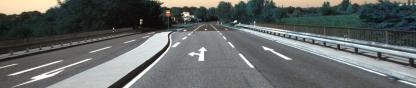}
    \\
    \rotatebox[origin=l]{90}{\,Baseline}
    &
    \includegraphics[width=\sfmwidth]{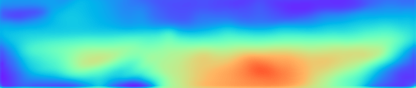}
    \\
    \rotatebox[origin=l]{90}{\,\,\,\,\,\,Ours}
    &
    \includegraphics[width=\sfmwidth]{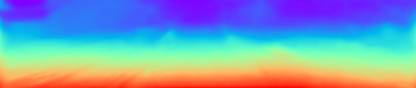}
    \\
    \rotatebox[origin=l]{90}{\,\,\,\,Truth}
    &
    \includegraphics[width=\sfmwidth]{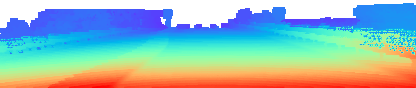}
    \\
  \end{tabular}
  \hspace{0.25in}
  \renewcommand{\nametag}{271}
  \begin{tabular}{@{}l@{\hspace{0.04in}}c@{}}
    \rotatebox[origin=l]{90}{\,\,\,\,\,\,Input}
    &
    \includegraphics[width=\sfmwidth]{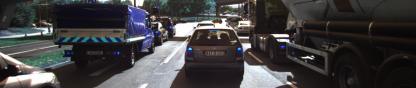}
    \\
    \rotatebox[origin=l]{90}{\,Baseline}
    &
    \includegraphics[width=\sfmwidth]{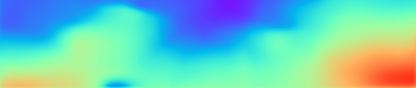}
    \\
    \rotatebox[origin=l]{90}{\,\,\,\,\,\,Ours}
    &
    \includegraphics[width=\sfmwidth]{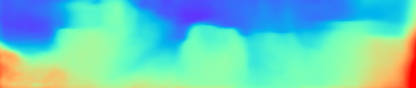}
    \\
    \rotatebox[origin=l]{90}{\,\,\,\,Truth}
    &
    \includegraphics[width=\sfmwidth]{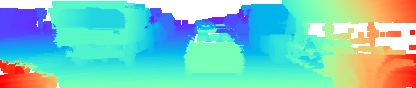}
    \\
  \end{tabular}
    \\
  \vspace{0.2in}
  \renewcommand{\nametag}{344}
  \begin{tabular}{@{}l@{\hspace{0.04in}}c@{}}
    \rotatebox[origin=l]{90}{\,\,\,\,\,\,Input}
    &
    \includegraphics[width=\sfmwidth]{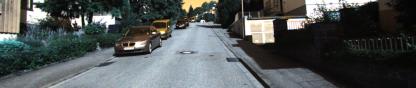}
    \\
    \rotatebox[origin=l]{90}{\,Baseline}
    &
    \includegraphics[width=\sfmwidth]{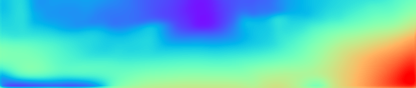}
    \\
    \rotatebox[origin=l]{90}{\,\,\,\,\,\,Ours}
    &
    \includegraphics[width=\sfmwidth]{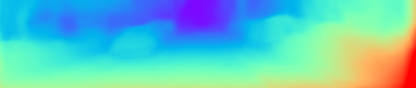}
    \\
    \rotatebox[origin=l]{90}{\,\,\,\,Truth}
    &
    \includegraphics[width=\sfmwidth]{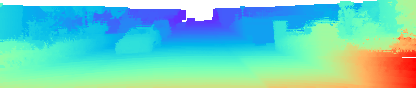}
    \\
  \end{tabular}
  \hspace{0.25in}
  \renewcommand{\nametag}{348}
  \begin{tabular}{@{}l@{\hspace{0.04in}}c@{}}
    \rotatebox[origin=l]{90}{\,\,\,\,\,\,Input}
    &
    \includegraphics[width=\sfmwidth]{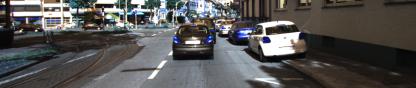}
    \\
    \rotatebox[origin=l]{90}{\,Baseline}
    &
    \includegraphics[width=\sfmwidth]{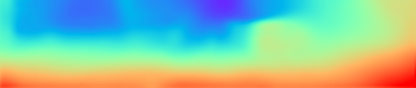}
    \\
    \rotatebox[origin=l]{90}{\,\,\,\,\,\,Ours}
    &
    \includegraphics[width=\sfmwidth]{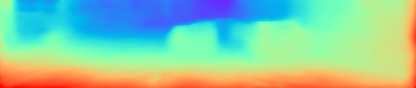}
    \\
    \rotatebox[origin=l]{90}{\,\,\,\,Truth}
    &
    \includegraphics[width=\sfmwidth]{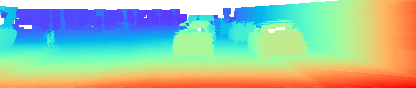}
    \\
  \end{tabular}
  \caption{
  Monocular depth estimation results on the KITTI benchmark using the ``Baseline'' network of \cite{zhou2017}
  and our own variant in which we replace the network's loss function with our own adaptive loss over wavelet coefficients.
  Changing only the loss function results in significantly improved depth estimates.
  \label{fig:sfm_results_big1}
  }
\end{figure*}

\begin{figure*}[p]
\centering
  \renewcommand{\nametag}{453}
  \begin{tabular}{@{}l@{\hspace{0.04in}}c@{}}
    \rotatebox[origin=l]{90}{\,\,\,\,\,\,Input}
    &
    \includegraphics[width=\sfmwidth]{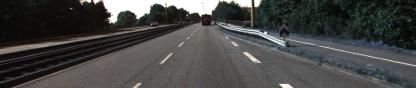}
    \\
    \rotatebox[origin=l]{90}{\,Baseline}
    &
    \includegraphics[width=\sfmwidth]{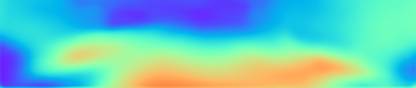}
    \\
    \rotatebox[origin=l]{90}{\,\,\,\,\,\,Ours}
    &
    \includegraphics[width=\sfmwidth]{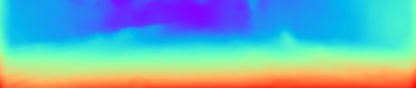}
    \\
    \rotatebox[origin=l]{90}{\,\,\,\,Truth}
    &
    \includegraphics[width=\sfmwidth]{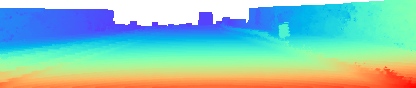}
    \\
  \end{tabular}
  \hspace{0.15in}
  \renewcommand{\nametag}{496}
  \begin{tabular}{@{}l@{\hspace{0.04in}}c@{}}
    \rotatebox[origin=l]{90}{\,\,\,\,\,\,Input}
    &
    \includegraphics[width=\sfmwidth]{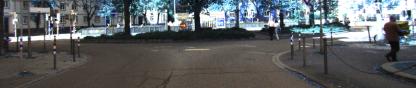}
    \\
    \rotatebox[origin=l]{90}{\,Baseline}
    &
    \includegraphics[width=\sfmwidth]{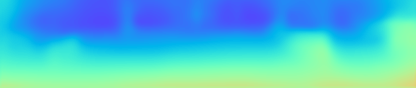}
    \\
    \rotatebox[origin=l]{90}{\,\,\,\,\,\,Ours}
    &
    \includegraphics[width=\sfmwidth]{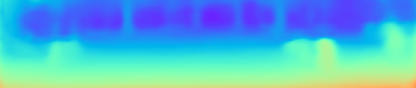}
    \\
    \rotatebox[origin=l]{90}{\,\,\,\,Truth}
    &
    \includegraphics[width=\sfmwidth]{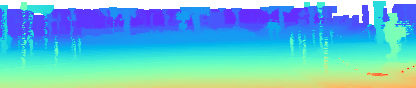}
    \\
  \end{tabular}
  \\
  \vspace{0.2in}
  \renewcommand{\nametag}{512}
  \begin{tabular}{@{}l@{\hspace{0.04in}}c@{}}
    \rotatebox[origin=l]{90}{\,\,\,\,\,\,Input}
    &
    \includegraphics[width=\sfmwidth]{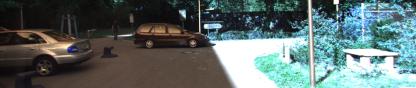}
    \\
    \rotatebox[origin=l]{90}{\,Baseline}
    &
    \includegraphics[width=\sfmwidth]{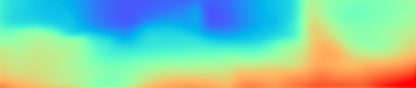}
    \\
    \rotatebox[origin=l]{90}{\,\,\,\,\,\,Ours}
    &
    \includegraphics[width=\sfmwidth]{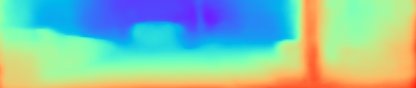}
    \\
    \rotatebox[origin=l]{90}{\,\,\,\,Truth}
    &
    \includegraphics[width=\sfmwidth]{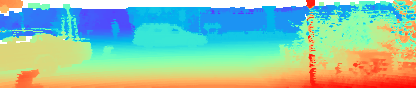}
    \\
  \end{tabular}
  \hspace{0.25in}
  \renewcommand{\nametag}{543}
  \begin{tabular}{@{}l@{\hspace{0.04in}}c@{}}
    \rotatebox[origin=l]{90}{\,\,\,\,\,\,Input}
    &
    \includegraphics[width=\sfmwidth]{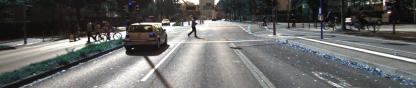}
    \\
    \rotatebox[origin=l]{90}{\,Baseline}
    &
    \includegraphics[width=\sfmwidth]{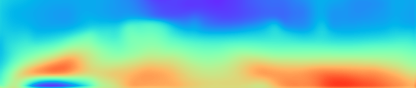}
    \\
    \rotatebox[origin=l]{90}{\,\,\,\,\,\,Ours}
    &
    \includegraphics[width=\sfmwidth]{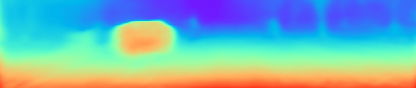}
    \\
    \rotatebox[origin=l]{90}{\,\,\,\,Truth}
    &
    \includegraphics[width=\sfmwidth]{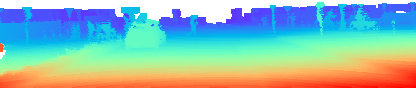}
    \\
  \end{tabular}
    \\
  \vspace{0.2in}
  \renewcommand{\nametag}{558}
  \begin{tabular}{@{}l@{\hspace{0.04in}}c@{}}
    \rotatebox[origin=l]{90}{\,\,\,\,\,\,Input}
    &
    \includegraphics[width=\sfmwidth]{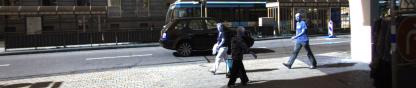}
    \\
    \rotatebox[origin=l]{90}{\,Baseline}
    &
    \includegraphics[width=\sfmwidth]{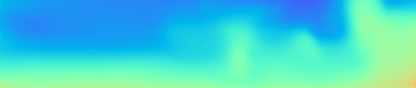}
    \\
    \rotatebox[origin=l]{90}{\,\,\,\,\,\,Ours}
    &
    \includegraphics[width=\sfmwidth]{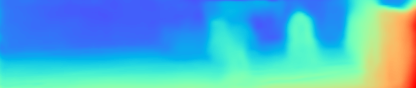}
    \\
    \rotatebox[origin=l]{90}{\,\,\,\,Truth}
    &
    \includegraphics[width=\sfmwidth]{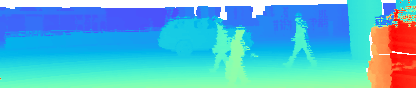}
    \\
  \end{tabular}
  \hspace{0.25in}
  \renewcommand{\nametag}{679}
  \begin{tabular}{@{}l@{\hspace{0.04in}}c@{}}
    \rotatebox[origin=l]{90}{\,\,\,\,\,\,Input}
    &
    \includegraphics[width=\sfmwidth]{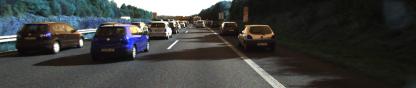}
    \\
    \rotatebox[origin=l]{90}{\,Baseline}
    &
    \includegraphics[width=\sfmwidth]{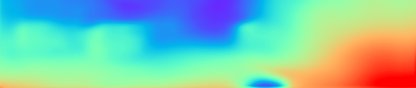}
    \\
    \rotatebox[origin=l]{90}{\,\,\,\,\,\,Ours}
    &
    \includegraphics[width=\sfmwidth]{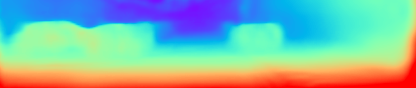}
    \\
    \rotatebox[origin=l]{90}{\,\,\,\,Truth}
    &
    \includegraphics[width=\sfmwidth]{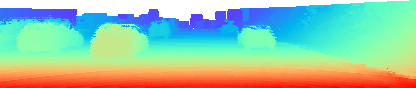}
    \\
  \end{tabular}
  \caption{
  Additional monocular depth estimation results, in the same format as
  Figure~\ref{fig:sfm_results_big1}.
  \label{fig:sfm_results_big2}
  }
\end{figure*}

\end{appendices}

\end{document}